\documentclass{article}

\usepackage[english]{babel}

\usepackage[letterpaper,top=2cm,bottom=2cm,left=3cm,right=3cm,marginparwidth=1.75cm]{geometry}

\usepackage{amssymb,amsmath,amsthm}
\usepackage{graphicx}
\usepackage[colorlinks=true, allcolors=blue]{hyperref}
\usepackage{breqn}

\usepackage{caption}
\usepackage{subcaption}

\usepackage{natbib}
\title{A Theory on Adam Instability in Large-Scale Machine Learning}
\author{Igor Molybog\footnote{\texttt{igormolybog@meta.com}}, Peter Albert, Moya Chen, Zachary DeVito,  \\ David Esiobu, Naman Goyal, Punit Singh Koura, Sharan Narang,  \\ Andrew Poulton, Ruan Silva, Binh Tang, Diana Liskovich, Puxin Xu, Yuchen Zhang, \\ Melanie Kambadur, Stephen Roller, Susan Zhang}
\date{%
    Meta AI \\
    \today
}

\begin{document}
\maketitle

\begin{abstract}
We present a theory for the previously unexplained divergent behavior noticed in the training of large language models. We argue that the phenomenon is an artifact of the dominant optimization algorithm used for training, called Adam. We observe that Adam can enter a state in which the parameter update vector has a relatively large norm and is essentially uncorrelated with the direction of descent on the training loss landscape, leading to divergence. This artifact is more likely to be observed in the training of a deep model with a large batch size, which is the typical setting of large-scale language model training. To argue the theory, we present observations from the training runs of the language models of different scales: 7 billion, 30 billion, 65 billion, and 546 billion parameters.

\end{abstract}


\section{Introduction}

Training instability reported by \citet{chowdhery2022palm} is an interesting phenomenon that has only been reported for the large language models trained on an order of a trillion tokens, posing a threat to further scaling of the AI systems. \citet{chowdhery2022palm} have observed dozens of spikes in the loss curve throughout training. To mitigate the issue, they re-started training from a checkpoint roughly 100 steps before the spike started, and skipped roughly 200–500 data batches, in order to exclude batches that were seen right before and during the spike. In that case, the spike of the loss value did not repeat. The spikes were also not observed when the skipped data was fed through the model again \textit{after} the aforementioned mitigation, which implies that the data itself did not cause the spike, but rather an interference of the data batch with the state of the model training run. The purpose of this work is to rigorously reproduce the experiment with a different hardware and software setup, come up with an explanation for the observed behavior supported by empirical evidence and theoretical arguments, and propose alternative ways of mitigating the issue.

Loss spikes are difficult to study because any reproduction of these spikes at a smaller scale is not necessarily caused by or remediated by the same factors as in larger scales.  We therefore analyze large-scale language modeling experiments, training four models between 7 billion and 546 billion parameters. The models are decoder-only transformers \citep{brown2020language, smith2022using} with different depth and embedding dimensions and trained using the AdamW \citep{loshchilov2017decoupled} algorithm with a linear learning rate schedule. Comparing to the modified Adafactor \citep{shazeer2018adafactor} used by \citet{chowdhery2022palm}, we did not use the "parameter scaling", $\beta_2$ build-up or the dynamic weight decay. This did not critically change the observed training instabilities.
We also made modifications in the architecture relative to the setup of \citet{chowdhery2022palm} so that the phenomenon we reproduce are robust to some changes in the specifics of model architectures. For example, we used the ReLU activation function like \citet{zhang2022opt} instead of SwiGLU \citet{shazeer2020}, and absolute learned positional embeddings instead of RoPE \citet{rope2021}. The settings of each training run that are important in the context of our analysis are displayed in Table \ref{tab:models}.
We cross-checked our results with the models trained using a different codebase and a different dataset, similar to those that were used for LLaMa \citep{touvron2023llama}, and did not see significant differences in our results at the scale of the $65$b model.

	\begin{table}[t]
\centering
\begin{tabular}{||c| c c c c c c c||}
 \hline
 model & depth & embedding dimension & $b$ (batch size) & $\eta_t$ (learning rate) & $\varepsilon$ & $\beta_1$ & $\beta_2$ \\ [0.5ex]
 \hline\hline
 $7$b & $32$ & $4096$ & $2048$ & $\approx 10^{-4}$ & $10^{-8}$ & $0.9$ & $0.95$ \\
 \hline
 $30$b & $36$ & $8192$ & $8192$ &$\approx 10^{-4}$ & $10^{-8}$ & $0.9$ & $0.95$ \\
 \hline
 $65$b & $80$ & $8192$ & $8192$ &$\approx 6\times 10^{-5}$ & $10^{-8}$ & $0.9$ & $0.95$ \\
 \hline
 $546$b & $108$ & $20480$ & $65536$ & $\approx 2\times 10^{-5}$ & $10^{-8}$ & $0.9$ & $0.95$ \\
 \hline
\end{tabular}
\caption{Training run settings}
\label{tab:models}
\end{table}

\begin{figure}
     \centering
     \begin{subfigure}[b]{1\textwidth}
         \centering
         \includegraphics[width=\textwidth]{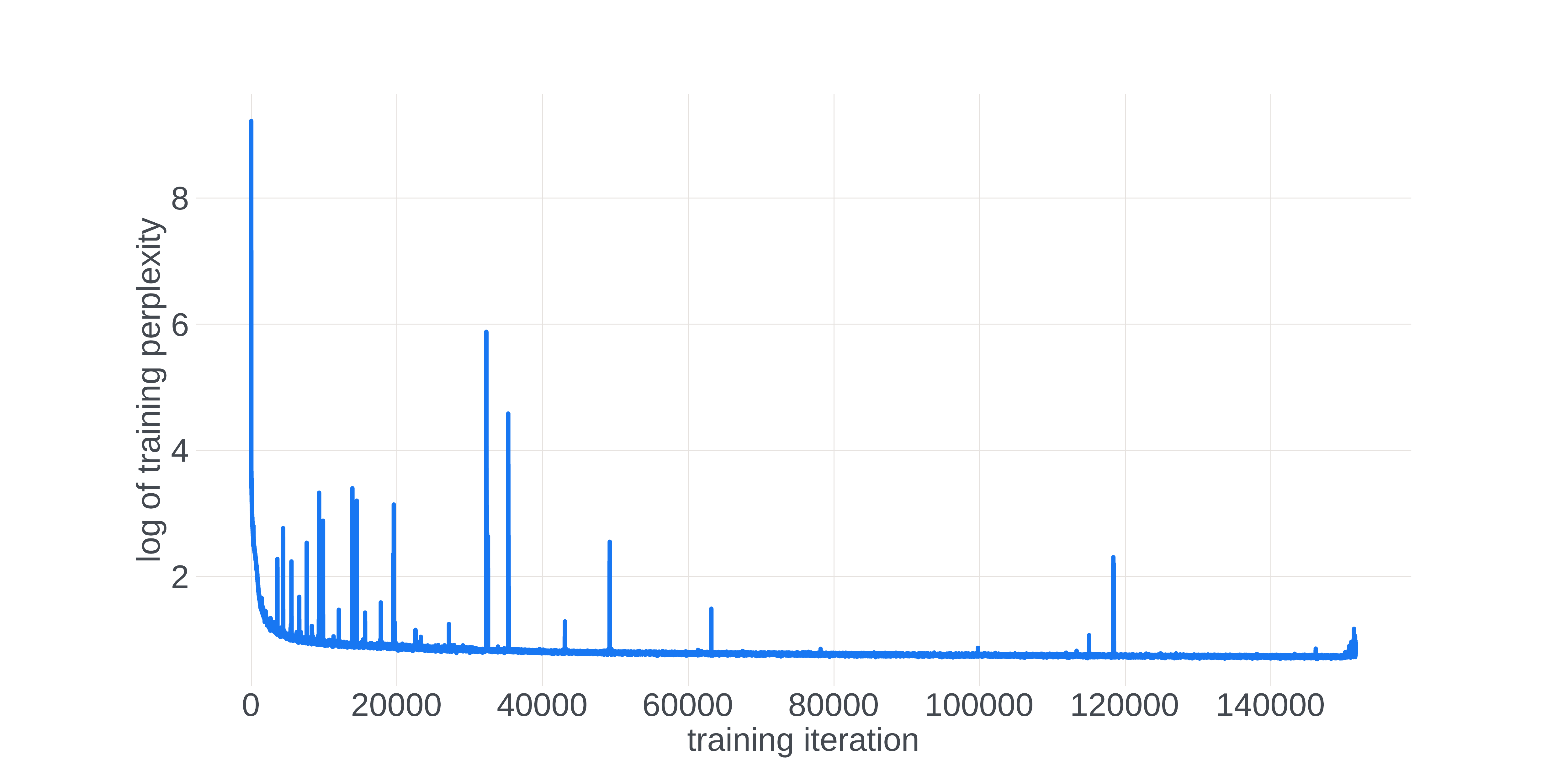}
     \end{subfigure}
        \caption{Training perplexity curve of $546$b model with prominent spikes}
        \label{fig:entire_curve}
\end{figure}

Training on a cluster of GPUs (while \citet{chowdhery2022palm} were working on TPUs), using a completely different codebase and datasets, we replicated the unstable behavior of the loss curve at the largest scale, which is demonstrated in Figure \ref{fig:entire_curve}. Although the batch skipping trick described earlier was also implemented for the reported training run, from most of the loss spikes the curve recovered without a batch skipping intervention and within an order of a dozen of training iterations. The spikes that had to employ the batch skipping trick are not displayed in Figure \ref{fig:entire_curve}. Our theory further also addresses the observation of loss curve recovery, besides the relationship between the training state and the data batch during the loss explosion.
The training run for the $65$b model experienced moderate instabilities compared to the $546$b, while training of the smaller models ($30$b and $7$b) did not show explosive divergence behavior at all.
Although we observe that the severity of instability depends on the data and architecture choice, we attribute its nature to the Adam algorithm itself as the only common element of all of the large-scale training experiments conducted up to date. To confirm or refute this point, however, additional empirical observations of a different learning algorithm, for example, Stochastic Gradient Descent (SGD) \citep{kiefer1952stochastic}, should be made in a similar setup.

The rest of the paper is structured as follows. In Section \ref{sec:prelim} we give the introduction on the notions considered and the key notation throughout the paper. In Section \ref{sec:adam} we elaborate on the specific properties of the Adam algorithm that we found relevant to explaining the loss spike behavior. Section \ref{sec:stat} contains theoretical prediction and empirical confirmation of statistical properties of the update rule of the Adam optimization algorithm, while Section \ref{sec:alignment} argues for the malicious nature of these properties.
Section \ref{sec:theory} lays out the step-by-step explanation of what is happening to the training run state during the loss spike. We further discuss the implications of our theory in Section \ref{sec:discussion}. Section \ref{sec:background} describes the most relevant background on the research devoted to the divergent behavior of Adam. Section \ref{sec:conclusion} provides a conclusion for our work.

\section{Prerequisites}\label{sec:prelim}

\subsection{Notations}

We interchangeably use the notation for matrices and vectors by referring to their components. An entry $i$ of a vector $x\in \mathbb R^n$ is denoted with $x[i]\in \mathbb R.$ For a subset of indices $G\subset \{1,\ldots,n\}$ of size $|G|,$ the vector obtained by combining together the entries of $x$ corresponding to $i\in G$ is denoted with $x[G] \in \mathbb R^{|G|}.$ For a matrix $x\in \mathbb R^{n\times T},$ the entry of row $i\in \{1,\ldots,n\}$ and column $t\in \{1,\ldots,T\}$ is denoted with $x[i,t]\in \mathbb R,$ while the $t$-th column is denoted with $x[:, t]\in \mathbb R^n$ or $x_t\in \mathbb R^n.$ Similarly, $x[G, t]\in \mathbb R^{|G|}$ denotes the sub-vector of $x[:, t]$ that corresponds to the subset $G\subset \{1,\ldots,n\}.$ For two tensors $x$ and $y,$ let $x \otimes y$ denote the outer product between $x$ and $y$ and $\langle x,y \rangle$ denote the standard inner product between them. The arithmetic operations on vectors, such as logarithm, square, and square root are implied to be coordinate-wise operations. We refer to the random eigenvalue of a constant matrix $x$ as $\lambda[x]$ (subject to uniform distribution over the eigenvalues).
The notation $x \sim D$ where $x$ is a random variable and $D$ is a distribution or a random variable means that the distributions of the two coincide almost surely.
If $x_t$ is an infinite sequence of random variables that is indexed with $t,$ then
$x_t \overset{d}{\to } D$ denotes the convergence of $x_t$ to $D$ in distribution.
The operators of the mean and variance of a random variable, mapping it to a scalar value, are denoted with $\mathbb E$ and $\mathbb D,$ respectively.
$\mathcal{N}(0, 1)$ denotes the standard Normal distribution. $\textrm{Bernoulli}(p)$ is the Bernoulli distribution that assigns the probability $p$ to the value of $1$ and $(1-p)$ to the value of $-1.$

\subsection{Adam algorithm}

Let us introduce the training procedure under investigation. It consists of a practical optimization algorithm that aims to minimize the empirical risk $f(\theta) = \mathbb E_{X\sim D} \ell_X(\theta)$ of a model $\theta$ subject to a dataset $D$ and a loss function $\ell$. The model parameters here are denoted with a vector of parameters $\theta\in \mathbb R^n.$ In the natural language processing context, the cross entropy loss function ties the empirical risk to the metric of perplexity, referred to in Figure \ref{fig:entire_curve}. In large language modeling practice, the optimization is performed using a variation of Adam \citep{kingma2014adam} optimization algorithm over a parametric family (architecture) of deep learning models called transformers \citep{vaswani2017attention}, following the choices made in the generative pre-trained transformer (GPT) line of research \citep{radford2019language, brown2020language}. The state-of-the-art models have the number of parameters of order of $100$ billion ($n\approx 10^{11}$) and the number of layers (depth) on the order of a hundred.

Adam is an iterative algorithm that uses first-order (gradient) information to update the model parameters $\theta_t$ on every iteration $t.$ It relies on the stochastic approximation $g_t$ of the gradient of the objective function $\nabla f(\theta_t),$ which is calculated as a gradient of the empirical risk subject to a portion of data samples called a batch. The number of data samples (size) $b$ of the batch is an important hyper-parameter of the algorithm that controls the variance of the gradient estimation $g_t.$ The Adam update rule can be written as follows:
\begin{align*}
 m_t &= \frac{\beta_1}{1-\beta_1^t} m_{t-1} + \frac{1-\beta_1}{1-\beta_1^t} g_t \\
 v_t &= \frac{\beta_2}{1-\beta_2^t} v_{t-1} + \frac{1-\beta_2}{1-\beta_2^t} g_t^2 \\
u_t &=\frac{m_t}{\sqrt{v_t} + \varepsilon} \\
 \theta_{t+1} &= \theta_t - \eta_t u_t
\end{align*}
where $\beta_1, \beta_2$ are averaging hyper-parameters, $\varepsilon$ is a stability hyper-parameter and $\eta_t$ is a step size (or a learning rate). The vectors $m_t$ and $v_t$ make up the optimizer state at the iteration $t$ and $u_t$ is called the (unscaled) update vector. The original Adam algorithm is the focus of our paper because it is simple and illustrative. Our analysis can be carried over to include the many modifications of the algorithm that are used in practice, including weight decay \citep{loshchilov2017decoupled}, model norm regularization and gradient clipping \citep{pascanu2012understanding}.

The values $m_t$ and $v_t$ are weighted average vectors of $g_\tau$ which could be calculated explicitly from the values of $g_{\tau}$ for $\tau\in \{1, \ldots, t\}.$ Let us define $w^{1}_t[\tau] := \frac{\beta_1^{t-\tau}(1-\beta_1)}{\prod_{T = \tau}^t(1-\beta_1^T)}$ and $w^{2}_t[\tau] := \frac{\beta_2^{t-\tau}(1-\beta_2)}{\prod_{T = \tau}^t(1-\beta_2^T)}$ which are the coefficients of $g[i,\tau]$ in the expressions for $m[i,t ]$ and $v[i,t]$ for any component $i\in\{1,\ldots, n\}:$
\[m[i,t ] = \sum_{\tau=1}^t w^{(1)}_t[\tau] g[i, \tau]\]
\[v[i,t] = \sum_{\tau=1}^t w^{(2)}_t[\tau] g[i, \tau]^2.\]
It can be shown that $\sum_{\tau=1}^t w_t^{(1)}[\tau] = \beta_1,$ $\sum_{\tau=1}^t w_t^{(2)}[\tau] = \beta_2$ and the sums $\sum_{\tau=1}^t (w_t^{(1)}[\tau])^2$ and $\sum_{\tau=1}^t (w_t^{(2)}[\tau])^2$ converge to finite values as $t\to \infty.$

\section{Assumptions underlying Adam efficacy}\label{sec:adam}
In this section, we explicitly go through the assumptions on the training setup that we believe give Adam the advantage over other first-order stochastic methods, such as SGD.
We point out the assumption of time-domain independence between the gradient estimations as the crucial assumption that has a high chance of being subtly violated along the optimization procedure.


Let us take a look at Taylor's expansion of the gradient of the loss in the proximity of a point $\theta^\star$:
\[\nabla f(\theta) = \nabla f(\theta^\star) + \nabla^2 f(\theta^\star)[\theta - \theta^\star] + o(\|\theta-\theta^\star\|).\]
After introducing the notation $\Delta = \theta-\theta^\star,$ the outer product of the gradient with itself takes the form
\begin{align*}
    \nabla f(\theta) \nabla f(\theta)^\top =&& \nabla f(\theta^\star)\nabla f(\theta^\star)^\top + \nabla^2 f(\theta^\star)\Delta f(\theta^\star)^\top + f(\theta^\star)\Delta ^\top \nabla^2 f(\theta^\star)^\top + \nabla^2 f(\theta^\star)\Delta \Delta ^\top \nabla^2 f(\theta^\star)^\top \\
    && + o(\|\Delta\|^2).
\end{align*}
If $\theta^\star$ is a first-order stationary point ($\nabla f (\theta^\star) = 0$):
\[\nabla f(\theta) \nabla f(\theta)^\top = \nabla^2 f(\theta^\star)\Delta \Delta ^\top \nabla^2 f(\theta^\star)^\top + o(\|\Delta\|^2).\]
Assuming $\theta$ is a random vector and taking the expectation with respect to the randomness in $\theta$ yields
\[\mathbb E_\Delta \nabla f(\theta) \nabla f(\theta)^\top = \mathbb E_\Delta\nabla^2 f(\theta^\star)\Delta \Delta ^\top \nabla^2 f(\theta^\star)^\top + \mathbb E_\Delta o(\|\Delta\|^2).\]
Due to the linearity of mathematical expectation,
\[\mathbb E_\Delta \nabla f(\theta) \nabla f(\theta)^\top = \nabla^2 f(\theta^\star)\mathbb E_\Delta \left[ \Delta \Delta ^\top \right ] \nabla^2 f(\theta^\star)^\top + \mathbb E_\Delta o(\|\Delta\|^2).\]
Consider an expansion of the covariance matrix of $\Delta$ into a scaled identity and a residual: $E_\Delta \left[ \Delta \Delta ^\top \right ] = \sigma^2 \mathbb I + \Sigma,$ then
\begin{align}\label{eq:Taylor_of_outer}
    \mathbb E_\theta \nabla f(\theta) \nabla f(\theta)^\top = \sigma^2 \nabla^2 f(\theta^\star) \nabla^2 f(\theta^\star)^\top + \nabla^2 f(\theta^\star)\Sigma \nabla^2 f(\theta^\star)^\top +\mathbb E_\Delta o(\|\Delta\|^2).
\end{align}

Due to the definition of $g_t$ in the Adam algorithm, $v_t$ can be viewed as an approximation of the diagonal of the matrix $\mathbb E_{\theta \sim \theta_\tau} \nabla f(\theta) \nabla f(\theta)^\top$ where $\theta_\tau$ is the distribution of the model weights over the latest several steps, the number of which is controlled by the value of $\beta_2.$ If the covariance matrix for this distribution can be decomposed into $\mathbb E_{\theta \sim \theta_\tau} \left[ (\theta - \theta^\star)(\theta - \theta^\star) ^\top \right ] = \sigma^2 \mathbb I + \Sigma$ such that $\sigma^2 \gg \|\Sigma\|,$ and $\sigma $ sufficiently small for the residual term $\mathbb E_\Delta o(\|\Delta\|^2)$ in Taylor's expansion \eqref{eq:Taylor_of_outer} to be negligible, then
\[v_t \approx \textrm{diag}\left (\mathbb E_{\theta \sim \theta_\tau} \nabla f(\theta) \nabla f(\theta)^\top\right ) \approx \sigma^2 \textrm{diag}\left ( \nabla^2 f(\theta^\star) \nabla^2 f(\theta^\star)^\top\right ).\]
Assuming that $\nabla^2 f(\theta^\star)$ is primarily diagonal, meaning that approximations like $\textrm{diag}(\nabla^2 f(\theta^\star) \nabla^2 f(\theta^\star)^\top) \approx \textrm{diag}(\nabla^2 f(\theta^\star))^2$ are valid, one could speculatively write that
\[\frac{1}{\sqrt{v_t}} \approx \sigma^{-1} \textrm{diag}(\nabla^2 f(\theta^\star))^{-1}\approx \sigma^{-1}\textrm{diag}(\nabla^2 f(\theta^\star)^{-1}). \]
Supposing additionally that the Hessian of the loss function is approximately constant over the support of $\theta_\tau,$ one could also make the approximation $\nabla^2 f(\theta_t) \approx \nabla^2 f(\theta^\star).$ This, together with the assumptions that $m_t \approx \nabla f(\theta_t),$ and $v_t \gg \varepsilon,$ would imply that the update of the Adam algorithm is an approximation of the update of the pure form of Newton's method, up to a constant multiplier:
\[u_t = \frac{m_t}{\sqrt{v_t} +\varepsilon} \approx \frac{m_t}{\sqrt{v_t}} \stackrel{\propto}{\approx} \left(\nabla^2 f (\theta_t)\right)^{-1} \nabla f(\theta_t).\]
The quality of such an approximation would impact the convergence of the Adam dynamics and could explain the empirical success and popularity of the Adam algorithm.

Note that even if $\nabla f(\theta^\star) \ne 0$ but $\|\nabla f(\theta^\star)\nabla f(\theta^\star)^\top\| \ll \sigma^2, $ the reasoning above goes through under the regularity assumption that $\mathbb E_\Delta \Delta = 0$ and slight modifications.

\section{The assumption of time-domain independence between gradient estimations}\label{sec:stat}

We further focus on the above assumption $\sigma^2 \gg \|\Sigma\|$  on the covariance matrix of the distribution of the model weights over the latest several steps in the training $\mathbb E_{\theta \sim \theta_\tau} \left[ (\theta - \theta^\star)(\theta - \theta^\star) ^\top \right ] = \sigma^2 \mathbb I + \Sigma$. We focus on this assumption because it can be violated on a relatively small time scale, which is in line with the short-term explosion phenomenon we study. One important implication of $\sigma^2 \gg \|\Sigma\|$ is that the non-diagonal components of the covariance matrix $\mathbb E_{\theta \sim \theta_\tau} \left[ (\theta - \theta^\star)(\theta - \theta^\star) ^\top \right ]$ are of a small magnitude.

One scenario in which $\mathbb E_{\theta \sim \theta_\tau} \left[ (\theta - \theta^\star)(\theta - \theta^\star) ^\top \right ]$ has small in magnitude off-diagonal entries is when the components of the update vector $u_t=\frac{m_t}{\sqrt{v_t} +\varepsilon}$ are independent over both time-domain and space-domain (over the space of model parameters). Projecting this requirement onto the gradient estimations, for any $i,j\in \{1,\ldots,n\},$ spanning the domain of model parameters (space-domain) and $t,$ $s,$ spanning the domain of algorithm iterations (time domain) such that $(i, t) \ne (j, s)$, we demand from the pair of scalar random variables $g[i,t],$ $g[j, s]$ (stochasticity comes from the data distribution) to be approximately independent (the joint distribution function is point-wise close to the product of individual distribution functions). We call this requirement the time-domain independence of gradient estimation components.
In this case, $(\theta_t - \theta^\star)$ becomes a vector of (almost) independent Markov chains (due to the property of independent increments), and the condition on its covariance matrix is guaranteed to hold.

\subsection{Manifestation of independent gradient estimates}\label{sec:stat_indep}

    In this section, we show how the independence of $g[i,t]$ can be detected from the model dynamics. For example, consider $g[i, \tau]\sim \textrm{Bernoulli} (\frac{1}{2})$ - i.i.d. for all $i\in \{1,\ldots,n\},$ $\tau \in \{1,\ldots,\infty\}$ and let $\varepsilon = 0.$
    \[u[i, t] = \frac{\sum_{\tau=1}^t w^{(1)}_t[\tau] g[i, \tau]}{\sqrt{\sum_{\tau=1}^t w^{(2)}_t[\tau] g[i, \tau]^2}}.\]
    Since $g[i, \tau]^2 = 1,$ and $\sum_{\tau=1}^t w_t^{(2)}[\tau] = \beta_2,$ it holds that $u[i, t] = \beta_2^{-\frac{1}{2}}\sum_{\tau=1}^t w^{(1)}_t[\tau] g[i, \tau].$ The random variables $w^{(1)}_t[\tau] g[i, \tau]$ are independent, with $\mathbb E w^{(1)}_t[\tau] g[i, \tau] = 0$ and $\mathbb E (w^{(1)}_t[\tau] g[i, \tau])^2 = (w^{(1)}_t[\tau])^2 = \mathbb D w^{(1)}_t[\tau] g[i, \tau].$ The sequence of partial sums $\Gamma_t(\beta_1) = \sum_{\tau=1}^t (w^{(1)}_t[\tau])^2$ converges and the finite limit we denote with $\lim_{t\to \infty}\Gamma_t(\beta_1) = \Gamma(\beta_1)=\Gamma_1.$
    By the Central Limit Theorem (Lyapunov Theorem, \citep{billingsley2008probability}), as $t \to \infty$
    \[\frac{1}{\Gamma_t(\beta_1)} \sum_{\tau=1}^t w^{(1)}_t[\tau] g[i, \tau]  \overset{d}{\to } \mathcal{N}(0, 1).\]

    Thus, for a large enough fixed value of $t,$ the distributions of the update vector components $u[:, t]$ under the assumption of independent Bernoulli $g[i,t],$ approach the distribution of $\frac{\Gamma(\beta_1)}{\sqrt{\beta_2} }G$ where $G$ is a standard Gaussian random variable.

    Note that the application of the Central Limit Theorem only requires $g[i, \tau]$ to be independent, and thus $g[i, \tau]^2 = 1$ is the only part of the reasoning that uses Bernoulli distribution of $g[i, \tau].$ Thus, the same line of argument could be used to argue that any distribution of $g[:, :]$ which is close to being composed of independent distributions of $g[i, \tau],$ would result in a bell-shaped distribution of $u[:, t]$ for a large enough fixed value of $t.$

\begin{figure}
     \centering
     \begin{subfigure}[b]{0.24\textwidth}
         \centering
         \includegraphics[width=\textwidth]{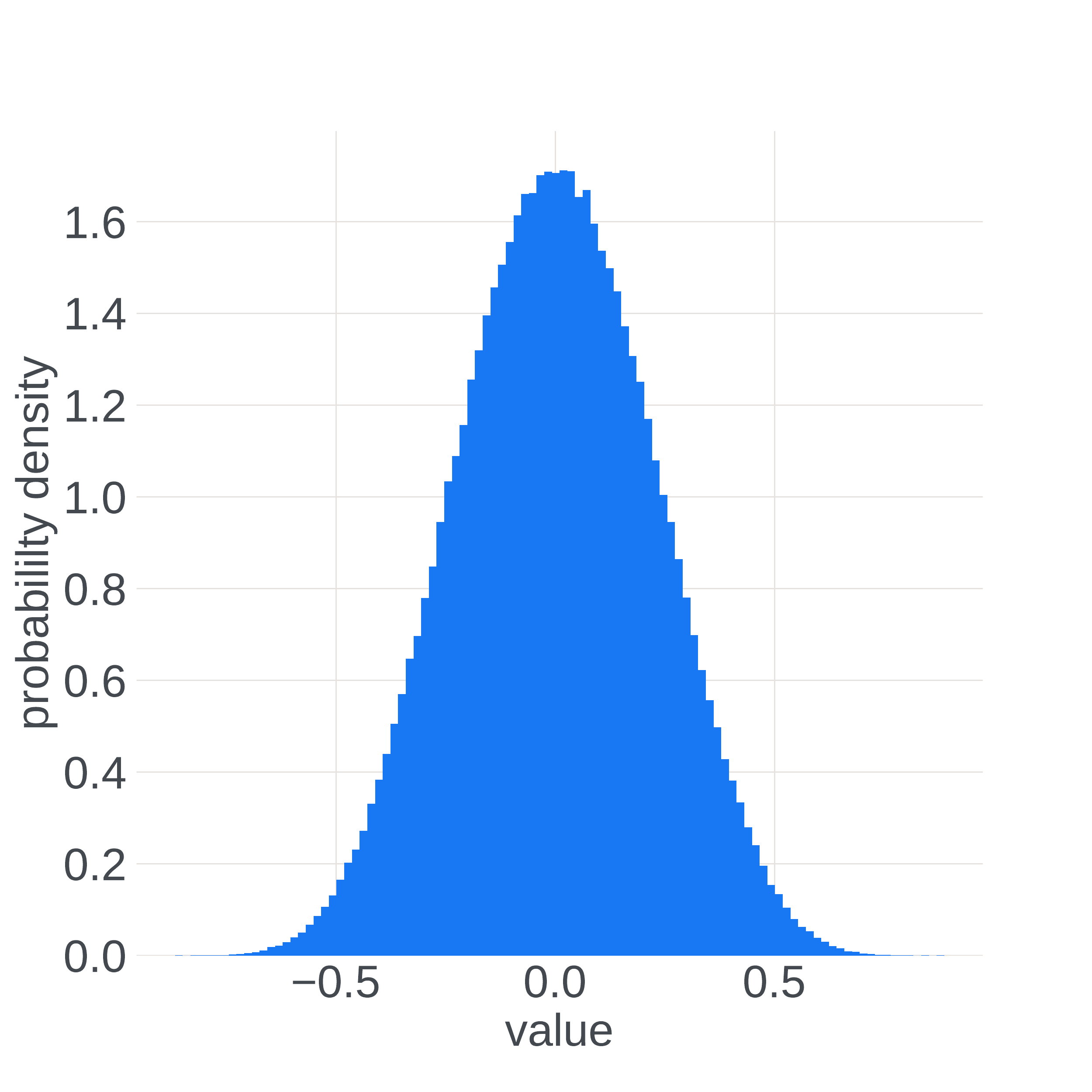}
         \caption{$7b$}
         \label{fig:bell7}
     \end{subfigure}
     \hfill
     \begin{subfigure}[b]{0.24\textwidth}
         \centering
         \includegraphics[width=\textwidth]{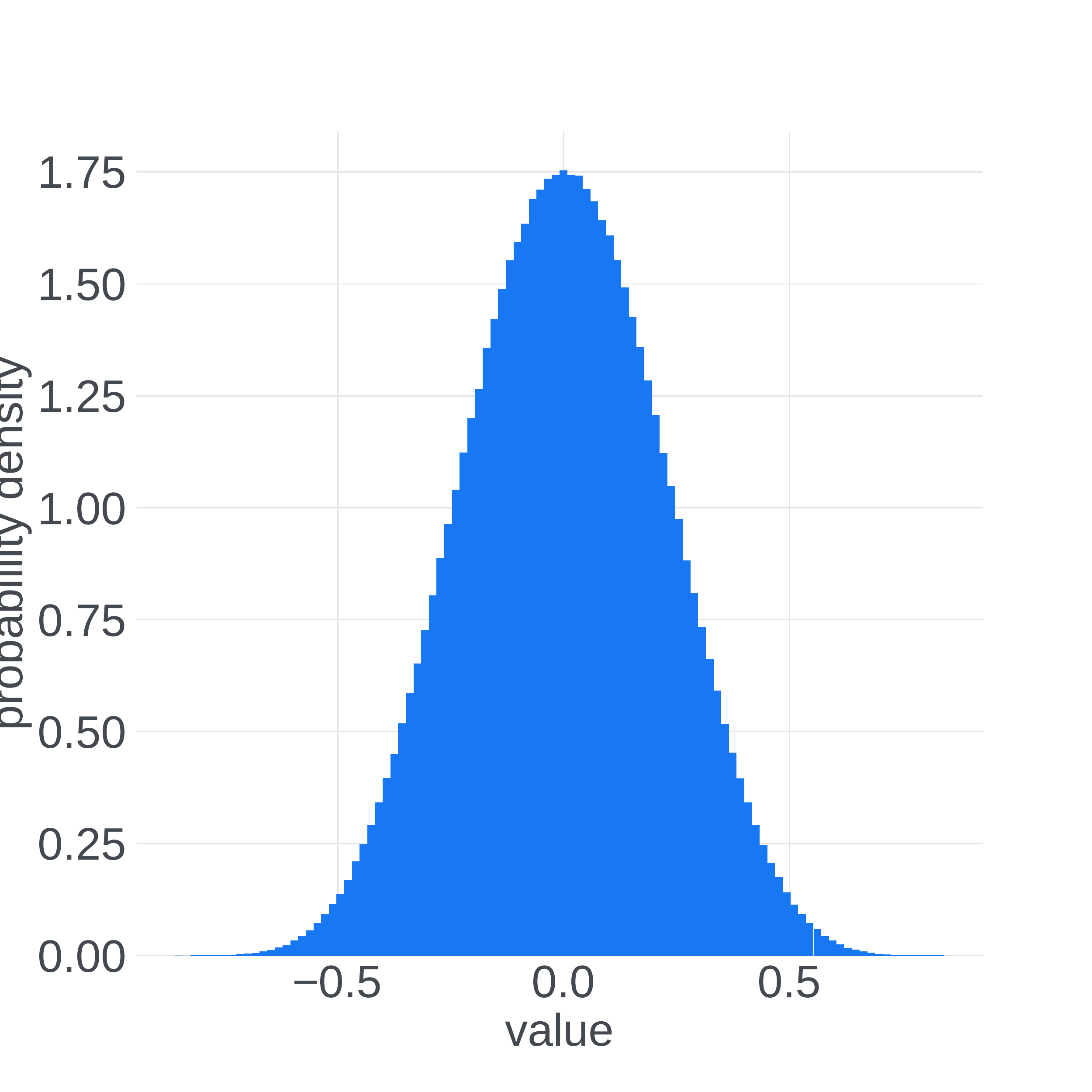}
         \caption{$30b$}
         \label{fig:bell30}
     \end{subfigure}
     \hfill
     \begin{subfigure}[b]{0.24\textwidth}
         \centering
         \includegraphics[width=\textwidth]{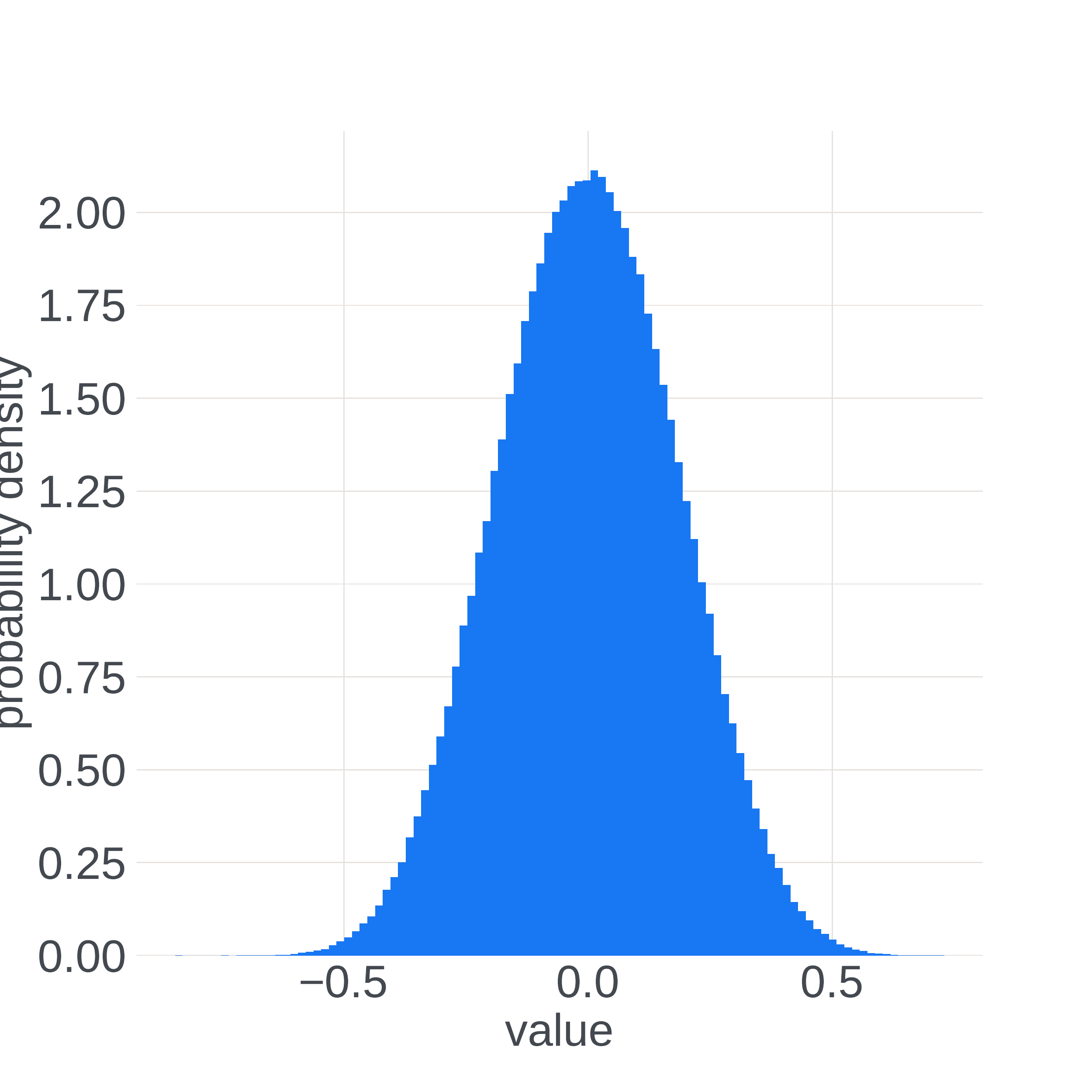}
         \caption{$65b$}
         \label{fig:bell65}
     \end{subfigure}
     \begin{subfigure}[b]{0.24\textwidth}
         \centering
         \includegraphics[width=\textwidth]{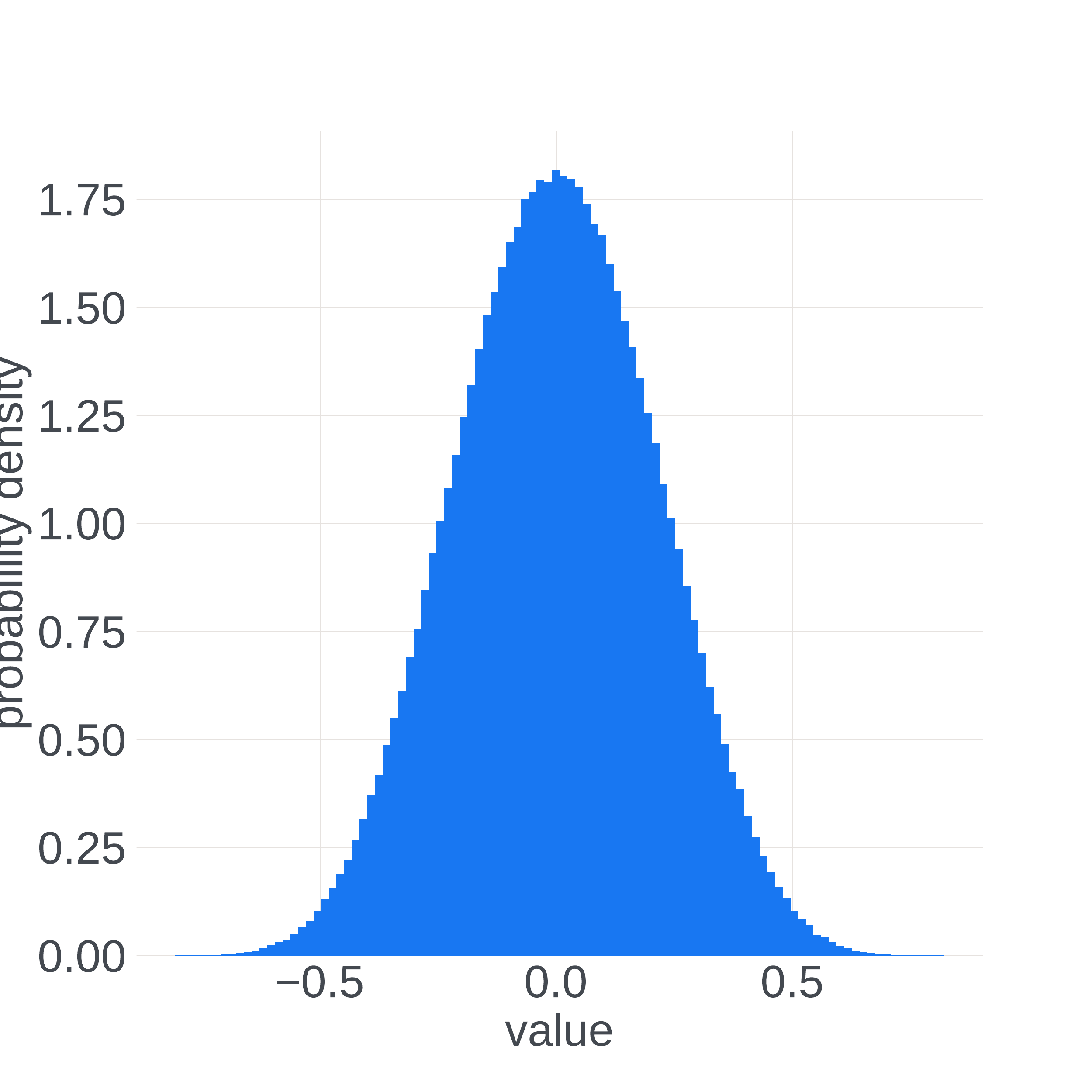}
         \caption{$546b$ ($t=133000$)}
         \label{fig:bell546}
     \end{subfigure}
        \caption{The distribution of $u[i, t]$ over the domain of model parameters $i\in G.$ Each plot corresponds to a layer $G$ of a model at the training iteration $t$ taken during the normal operation of the Adam algorithm, showing healthy converging dynamic.}
        \label{fig:bells}
\end{figure}

    To illustrate this point, we present examples of the distribution of the update $u[:, t]$ for time steps $t$ where the training loss is showing a healthy converging dynamic. Figure \ref{fig:bells} contains the probability density functions of $u[i, t]$ over $i\in G$ where $G$ is a group of model parameters that make up a layer of the network. It is remarkable how similar the distributions are for different models and different steps of the model training $t.$

\subsection{Manifestation of time-domain correlated gradient estimates}\label{sec:stat_dep}

    Now, let us see what the distribution of the update value across the model looks like if there is a time-domain correlation in the distribution of $g[i, t].$ Assume that $g[i, t] = g^{(1)}[i] + g^{(2)}[i, t] $ where the scaled Bernoulli variables $g^{(1)}[i]\sim \rho\times \textrm{Bernoulli}(\frac{1}{2})$ and $ g^{(2)}[i, t]\sim \textrm{Bernoulli}(\frac{1}{2})$ are independent random variables for all $i \in \{1,\ldots, n\}$ and $t\in\{1,\ldots,\infty\}$ . Note that $\mathbb E g[i, t] =0$ and
    \[\mathbb E g[i, t] =  \underbrace{\mathbb E g^{(1)} [i]^2}_{=1} + \underbrace{\mathbb E g^{(2)} [i, t]^2}_{\rho^2} + 2\underbrace{\mathbb E g^{(1)}[i] g^{(2)}[i, t]}_{=\mathbb E g^{(1)}[i] \mathbb E g^{(2)}[i, t]=0} = 1 + \rho^2 = \mathbb D g[i, t]. \]
    Under the assumption $\varepsilon = 0$ and using that $g[i, t]^2 = g^{(1)}[i]^2 + g^{(2)}[i, t]^2 + 2g^{(1)}[i] g^{(2)}[i, t] = 1+\rho^2+2g^{(1)}[i] g^{(2)}[i, t],$ we write the expression for a component of an Adam update vector:
    \[u[i,t] = \frac{\sum_{\tau=1}^t w^{(1)}_t[\tau] g[i, \tau]}{\sqrt{\sum_{\tau=1}^t w^{(2)}_t[\tau] g[i, \tau]^2}} = \frac{\sum_{\tau=1}^t w^{(1)}_t[\tau] g^{(2)}[i, \tau] + g^{(1)}[i]\overbrace{\sum_{\tau=1}^tw^{(1)}_t[\tau]}^{=\beta_1} }{\sqrt{(1+\rho^2)\underbrace{\sum_{\tau=1}^t w^{(2)}_t[\tau]}_{=\beta_2} + 2g^{(1)}[i] \sum_{\tau=1}^t w^{(2)}_t[\tau] g^{(2)}[i, \tau]}}.\]
    Again, by the Central Limit Theorem (Lyapunov Theorem, \citep{billingsley2008probability}),
    \begin{align*}
        \sum_{\tau=1}^t w^{(1)}_t[\tau] g^{(2)}[i, t] &\stackrel{d}{\to}\Gamma(\beta_1) G[i] = \Gamma_1 G[i] \\
       \sum_{\tau=1}^t w^{(2)}_t[\tau]g^{(2)}[i, t] &\stackrel{d}{\to} \Gamma(\beta_2) G[i] = \Gamma_2 G[i]
    \end{align*}
    where $G[i]$ are independent standard Gaussian random variables. Thus, for a large value of $t,$ the components of the update vector can be considered to come from the distribution
    \[u[i,t]\sim \frac{\Gamma_1 G[i] + \beta_1 g^{(1)}[i]}{\sqrt{\beta_2(1+\rho^2) + 2\Gamma_2 g^{(1)}[i] G[i] }}.\]
    We can distinguish two regimes for this distribution, depending on the value of $\rho$. If $\rho \ll 1,$ which corresponds to low time-domain correlation in $g[i,t],$ then the distribution takes a bell-shape similar to $\frac{\Gamma_1 G}{\sqrt{\beta_2  }}.$

    In the case of a very strong time-domain correlation $(\rho \gg 1),$ the distribution of $G$ can be considered supported in the proximity of $0,$ and thus the distribution of $u[i,t]$ takes the form close to $\frac{\beta_1 g^{(1)}[i]}{\rho\sqrt{\beta_2 }}$, which is a symmetrical bimodal distribution with the modes centered around $\pm \frac{\beta_1}{\sqrt{\beta_2}}.$ It is remarkable that neither of these distributions depends on the scale of the gradient values themselves. Again, due to the versatile Central Limit Theorem used here, the reasoning above could be applied to the distributions beyond Bernoulli. Thus, a similar bimodal picture of the distribution of the Adam update should be expected in the general case of correlated gradient estimations.

    Intuitively, this picture can be explained as follows: if for any fixed $i$ the values of $g[i, t]$ are the same for all $t$ ($g[i, t] = g[i]$ or $g_t = g$), then the vector of updates consists of the values $\pm \frac{\beta_1}{\sqrt{\beta_2}}$ as $u_t\big |_{\varepsilon = 0} = \textrm{sign}(g)\frac{\beta_1}{\sqrt{\beta_2}}$ for all $t.$

\begin{figure}
     \centering
     \begin{subfigure}[b]{0.24\textwidth}
         \centering
         \includegraphics[width=\textwidth]{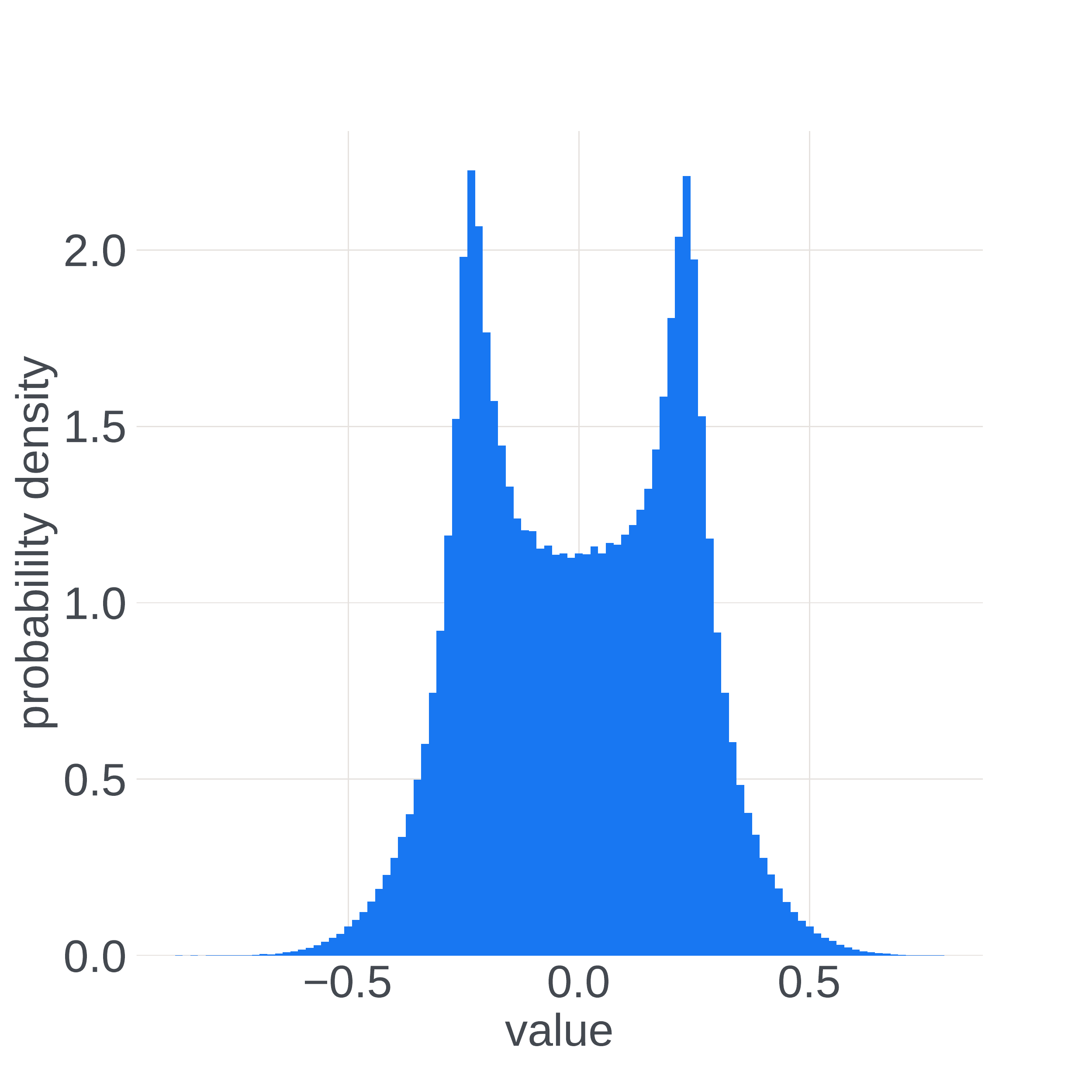}
         \caption{$65b$}
         \label{fig:bimodal_u65}
     \end{subfigure}
     \begin{subfigure}[b]{0.24\textwidth}
         \centering
         \includegraphics[width=\textwidth]{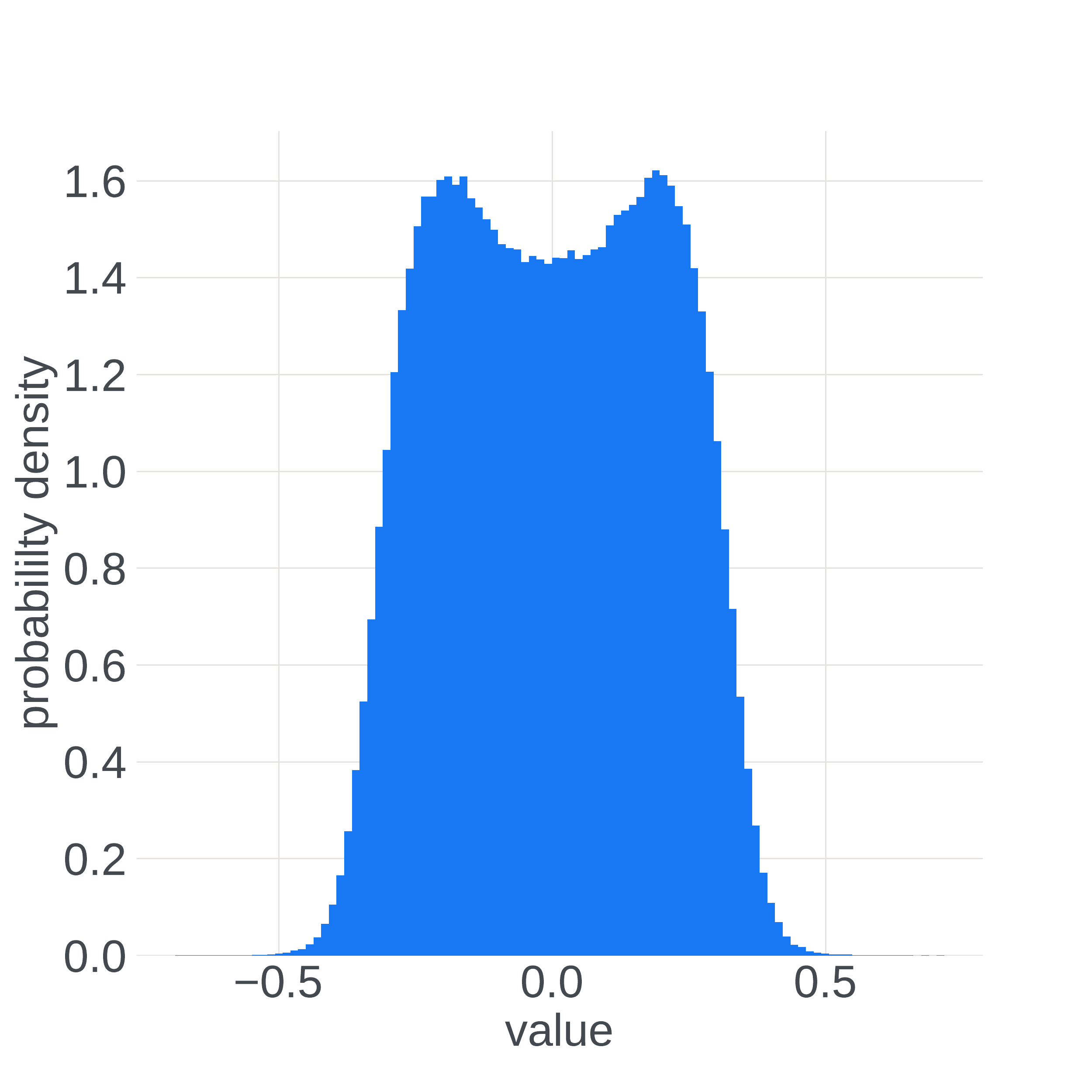}
         \caption{$546b$ ($t=149750$)}
         \label{fig:bimodal_u546}
     \end{subfigure}
        \caption{The distribution of $u[i, t]$ over the domain of model parameters $i\in G.$ Each plot corresponds to a layer $G$ of a model at the training iteration $t$ taken during the period of a brief loss stall in the training dynamic.}
        \label{fig:bimodal_u}
\end{figure}

    Anywhere between these two regimes ($\rho \approx 1$), one would observe a bimodal symmetrical distribution with a significant overlap between the modes.
    For example, we draw distributions of the update $u[:, t]$ for time steps $t$ when the training loss was stalling or showing divergent behavior. Figure \ref{fig:bimodal_u} contains the probability density functions of $u[i, t]$ over $i\in G$ where $G$ is a group of variables that make up a layer of the network. It is important to note that the majority of the layers in the model weight snapshots used to draw these images were still distributed close to the bell-shaped distribution depicted in Figure \ref{fig:bells}. Thus, a severe time-domain correlation between gradient estimations in a single layer should be associated with the convergence properties of the entire model. The number of the layers that have changed their update distribution from the bell-shaped distributions to the bimodal distributions has been increasing for our larger models as the training unrolled.
    Another important remark is that we were not able to detect layers with a bi-modal distribution of the updates in the smaller models that did not experience any perplexity spikes, namely $7$b and $30$b.

\section{
    Time-domain correlation leads to divergence of Adam}\label{sec:alignment}

In Section \ref{sec:adam} we have discussed why time-domain independence between gradient estimation components is important for fast convergence. Here, we continue by showing how its lack leads not only to slow convergence but to divergence of Adam iterations, which allows us to link it to the loss instability phenomenon. In this Section, we assume that the gradient estimates are exact but correlated and show that the learning rate required for convergence falls dramatically as $\frac{1}{n}.$ We do it by relying on the property that, in a high-dimensional space, a random vector is almost orthogonal to its sign.

     Let us look at Taylor's expansion of
    \[f(\theta_{t+1}) = f(\theta_{t} - \eta_t u_t) = f(\theta_t) - \eta_t \langle\nabla f(\theta_t), u_t\rangle + \frac{\eta_t^2}{2} \langle\nabla^2 f(\theta_t), u_t \otimes u_t \rangle + \ldots.\]
    The loss difference can be written as
    \[f(\theta_{t+1}) - f(\theta_t)= - \eta_t \langle\nabla f(\theta_t), u_t\rangle + \frac{\eta_t^2}{2} \langle\nabla^2 f(\theta_t), u_t \otimes u_t \rangle + o(\eta_t^2).\]
    Considering small values of the step size $\eta_t,$ the Adam step leads to a decreased loss value in case the following inequality holds:
    \begin{align}
    \label{eq:descent_condition}
    \langle\nabla f(\theta_t), u_t\rangle > \frac{\eta_t}{2} \langle\nabla^2 f(\theta_t), u_t \otimes u_t \rangle.
    \end{align}

Let us consider the ideal case when the gradient estimations are exact ($\nabla f(\theta_t) = g_t$ for all $t$), but there is a strong time-domain correlation between the gradient estimations ($\nabla f(\theta_t) = g_t = g $ for all $t$), which implies that $m_t = \beta_1\nabla f(\theta_t) $ and $v_t = \beta_2\nabla f(\theta_t)^2.$

\begin{itemize}
    \item Assume $\varepsilon \gg  \max_i\left|\frac{\partial f(\theta_t)}{\partial \theta_t[i]}\right|:$

    We can conclude that in this scenario $u_t = \frac{m_t}{\sqrt{v_t}+\varepsilon} = 0.$
    Both the left-hand side and right-hand side of the inequality \eqref{eq:descent_condition} are equal $0,$ the inequality is not satisfied.
    \item Assume $\varepsilon \ll \min_i\left|\frac{\partial f(\theta_t)}{\partial \theta_t[i]}\right|$ (e.g. $\varepsilon = 0$):

      \[u_t = \frac{\beta_1}{\sqrt{\beta_2}}\textrm{sign}(\nabla f(\theta_t))\]
     Divide the inequality \eqref{eq:descent_condition} by $\|u_t\|_2$ to obtain \[\frac{\langle\nabla f(\theta_t), u_t\rangle}{\langle u_t,u_t \rangle} > \frac{\eta_t}{2} \langle\nabla^2 f(\theta_t), \frac{u_t}{\|u_t\|} \otimes \frac{u_t}{\|u_t\|} \rangle. \]
     The left-hand side of this inequality is equal to $\frac{\beta_1}{\sqrt{\beta_2}}\frac{\|\nabla f(\theta_t)\|_1}{n}$ and the right-hand side is a random eigenvalue of the hessian of $f$ at $\theta_t$ scaled by $\eta_t/2.$ We should assume that it is an eigenvalue of a random order because there is no relationship between the direction of $u_t$ and the spectrum of the matrix $\nabla^2 f(\theta_t).$

     The spectrum of the hessian of a neural network loss function has been the focus of various studies \citep{sankar2021deeper, liao2021hessian, yao2020pyhessian}. None of them were considering the question of how the spectrum evolves as the model is being scaled up, and there is a signal that a rigorous study would be extremely difficult to conduct as \citet{liao2021hessian} pointed out that the spectrum can not be exactly described as a semicircular or even Marchenko-Pastur distribution already for simple generalized generalized linear models (GGLM). Thus, we will reason about the scaling of the average eigenvalue of a randomly generated matrix to provide some perspective on how the eigenvalues of the hessian could scale.

     According to Wigner's Semicircle law (see e.g. \citep{liu2000statistical}), a symmetric matrix with the upper-triangle part filled with i.i.d. random variables of zero mean, unit variance, and finite higher-order moments, has the distribution of normalized and shifted eigenvalue ($\frac{\lambda}{2\sqrt{n}} + \frac{1}{2}$) that converges to $\textrm{Beta}(\frac{3}{2}, \frac{3}{2})$ distribution as $n \to \infty$.
     The density of a normalized eigenvalue ($\frac{\lambda}{2\sqrt{n}}$) appears as a semicircle and has the explicit formula $\mathbb P(x) = \frac{2}{\pi}\sqrt{1-x^2}$. This distribution is symmetric with the center at $0$ and thus the expected value of the eigenvalue is equal to zero as well: $\mathbb E \lambda = 0.$ Throughout optimization, however, for $t$ large enough, one would expect $\theta_t$ to be in proximity of a locally optimal point, which, by the second-order necessary conditions of local optimality, means that the Hessian $\nabla^2 f(\theta_t)$ ought to be close to being a positive semi-definite matrix. Thus, modeling the Hessian as a matrix of random i.i.d. entries is not entirely correct. We can model it as a square of a random symmetric matrix that is guaranteed to be a positive semi-definite matrix. Since the spectrum of a matrix squared is the square of the spectrum of the matrix, we are interested in the mean value of the square $\lambda^2$ of the eigenvalues of a symmetric random matrix with i.i.d. values. The second moment of the $\textrm{Beta}(\frac{3}{2}, \frac{3}{2})$ distribution is equal to $\frac{5}{16},$
     which is also equal to the mean value of $(\frac{\lambda}{2\sqrt{n}} + \frac{1}{2})^2 = \frac{\lambda^2}{4n} + \frac{\lambda}{\sqrt{n}} +\frac{1}{4}.$ Since $\mathbb E \lambda = 0,$ we conclude that $\mathbb E \lambda^2 = \frac{n}{4}\propto n$ for a symmetric positive semi-definite matrix of size $n\times n$ with the entries of order $1,$ generated as the square of a symmetric matrix filled with i.i.d. entries.

    With this in mind, we can rewrite the condition on the learning rate stated by the inequality \eqref{eq:descent_condition} in the form
    \[\frac{2\beta_1}{n\sqrt{\beta_2}} > \eta_t \frac{\mathbb E \lambda\left[ \nabla^2 f(\theta_t) \right]}{\|\nabla f(\theta_t)\|_1}.\]
    If we assume that the entries of $\nabla f(\theta_t)$ and $\nabla^2 f(\theta_t)$ are of the same order of magnitude, then both the numerator and denominator of the right-hand side of this inequality should scale linearly with $n,$ thus requiring the learning rate $\eta_t$ to scale as $\frac{1}{n}$ to avoid divergence. This is not a realistic scale for the learning rate in the context of large-scale machine learning.

\end{itemize}

    Thus, we conclude that the requirement of time-domain incoherence between the gradient estimations is not only one of the necessary conditions for $\frac{1}{\sqrt{v_t}}$ to be an efficient estimation of the diagonal of the hessian inverse, as shown in Section \ref{sec:adam}, but also a necessary condition for convergence of the Adam algorithm in general.

\section{Theory of the loss instability}\label{sec:theory}

In the previous sections, we have shown that the time-domain independence of gradient estimation components is an important property for the convergence of the Adam algorithm. We have seen that it holds for the training runs that are showing stable behavior and can be violated in the periods of training instability. Further, we investigate the reasons why the time-domain independence of gradient estimates can be violated for some layers.

\begin{figure}
     \centering
     \begin{subfigure}[b]{1\textwidth}
         \centering
         \includegraphics[width=\textwidth]{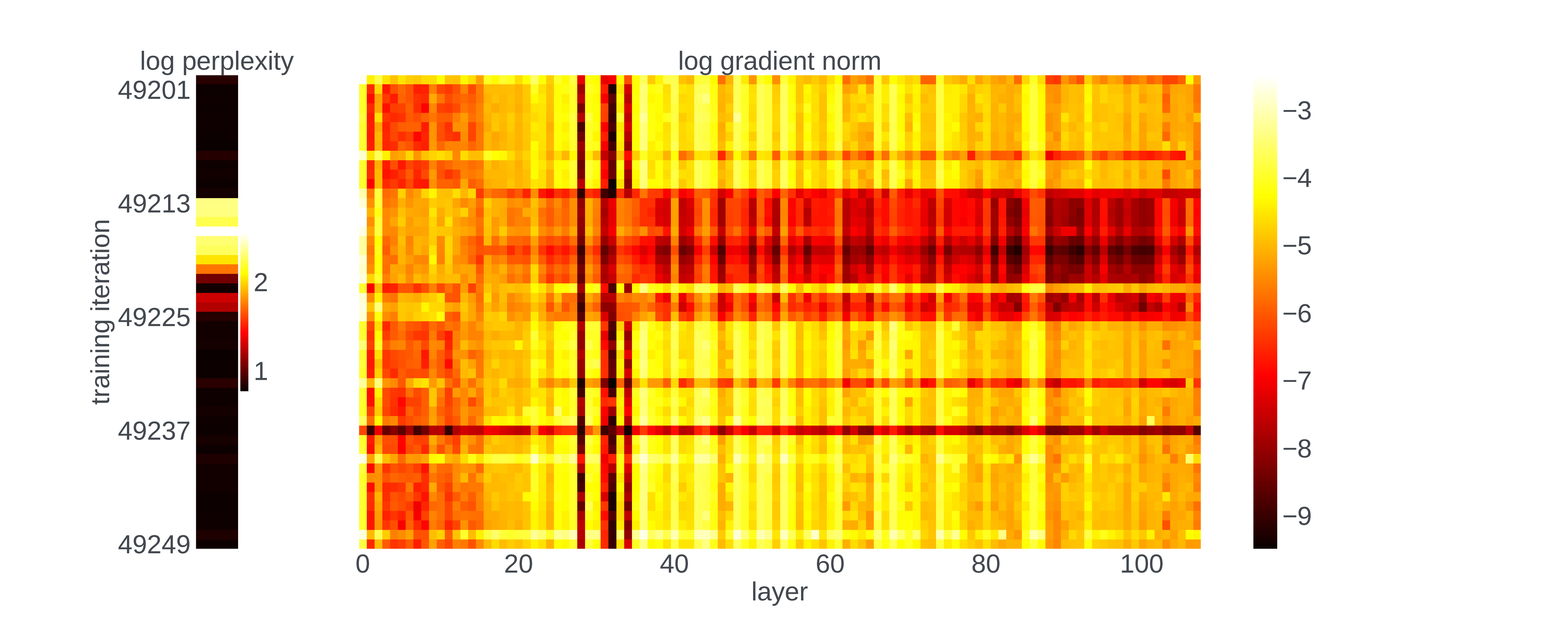}
         \caption{training iteration $t \approx 50000$}
         \label{fig:heatmap_50}
     \end{subfigure}
     \begin{subfigure}[b]{1\textwidth}
         \centering
         \includegraphics[width=\textwidth]{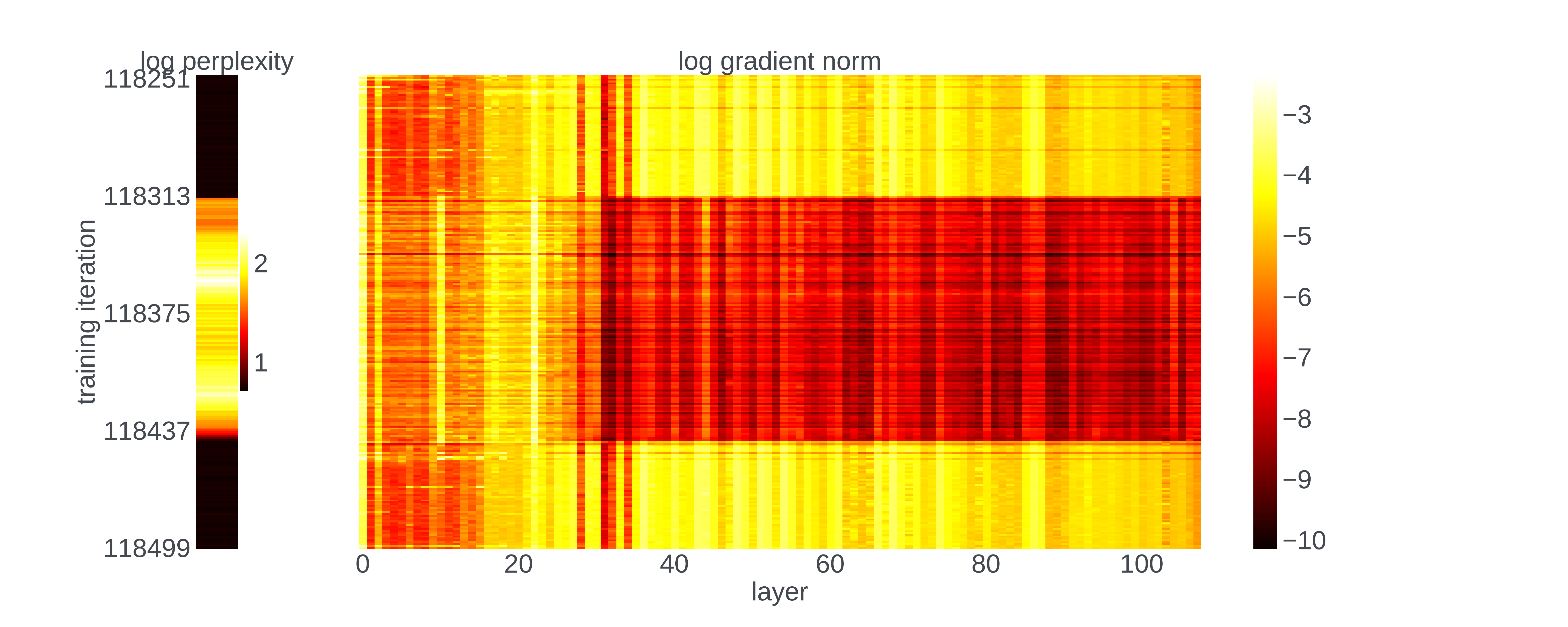}
         \caption{Training iteration $t \approx 120000$}
         \label{fig:heatmap_120}
     \end{subfigure}
        \caption{The pixels of the heat map to the right represent the log norm of the gradient estimation of a layer (x axis) at an iteration (y axis). The pixels of the strip to the left represent the values of the log perplexity at each iteration (y axis).}
        \label{fig:heatmaps}
\end{figure}

Let us turn to Figure \ref{fig:heatmaps}. There, we plot the behavior of the gradient norm and the perplexity in the proximity of two perplexity spikes observed in the model $546$b. In the earlier layers of the network, the gradient estimations get the norm that is orders of magnitude smaller than the layers further off in the model. Moreover, there are four layers that get the gradient estimates of a norm significantly below the value of $\varepsilon=10^{-8}.$ Right before the explosion starts, the norm of the gradient estimation increases for the earlier layers, and decreases for the rest. In the next iteration, the layers that were persistently getting the gradient estimation of a norm below $\varepsilon,$ experience an increased magnitude of the gradient estimation, and the perplexity value climb up. The gradient estimation norms of the earlier layers keep the high values throughout the loss explosion period and only scale back to the low values after the loss returns to its pre-explosion regime. Although not depicted on the Figure, the norm of the gradient estimation of the entire model significantly increases during the period of the loss spike, despite the majority of the layers getting gradient estimation of a smaller norm.

\begin{figure}
     \centering
     \begin{subfigure}[b]{0.24\textwidth}
         \centering
         \includegraphics[width=\textwidth]{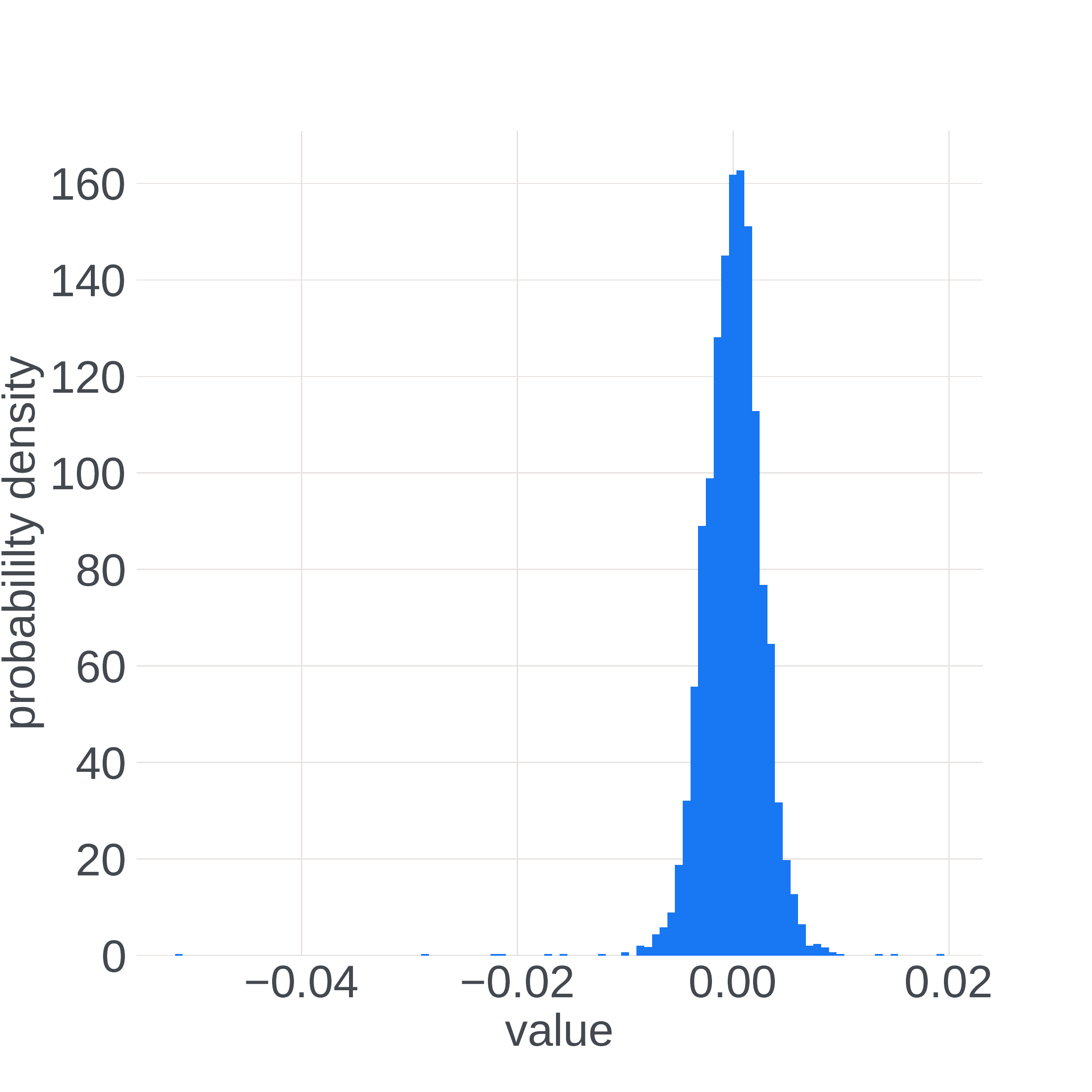}
         \caption{$6.7b$}
         \label{fig:spike_7b}
     \end{subfigure}
     \hfill
     \begin{subfigure}[b]{0.24\textwidth}
         \centering
         \includegraphics[width=\textwidth]{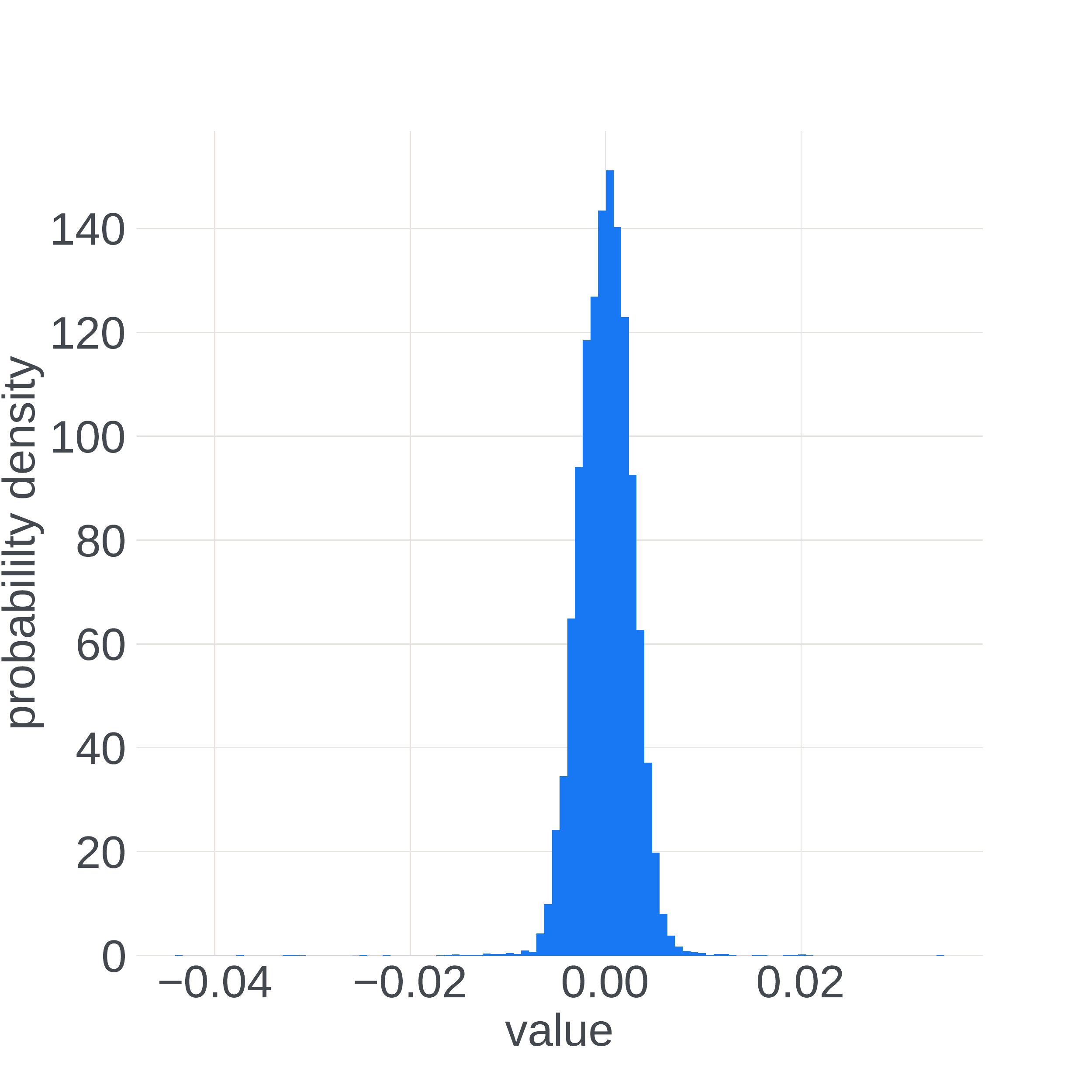}
         \caption{$30b$}
         \label{fig:spike_30b}
     \end{subfigure}
     \hfill
     \begin{subfigure}[b]{0.24\textwidth}
         \centering
         \includegraphics[width=\textwidth]{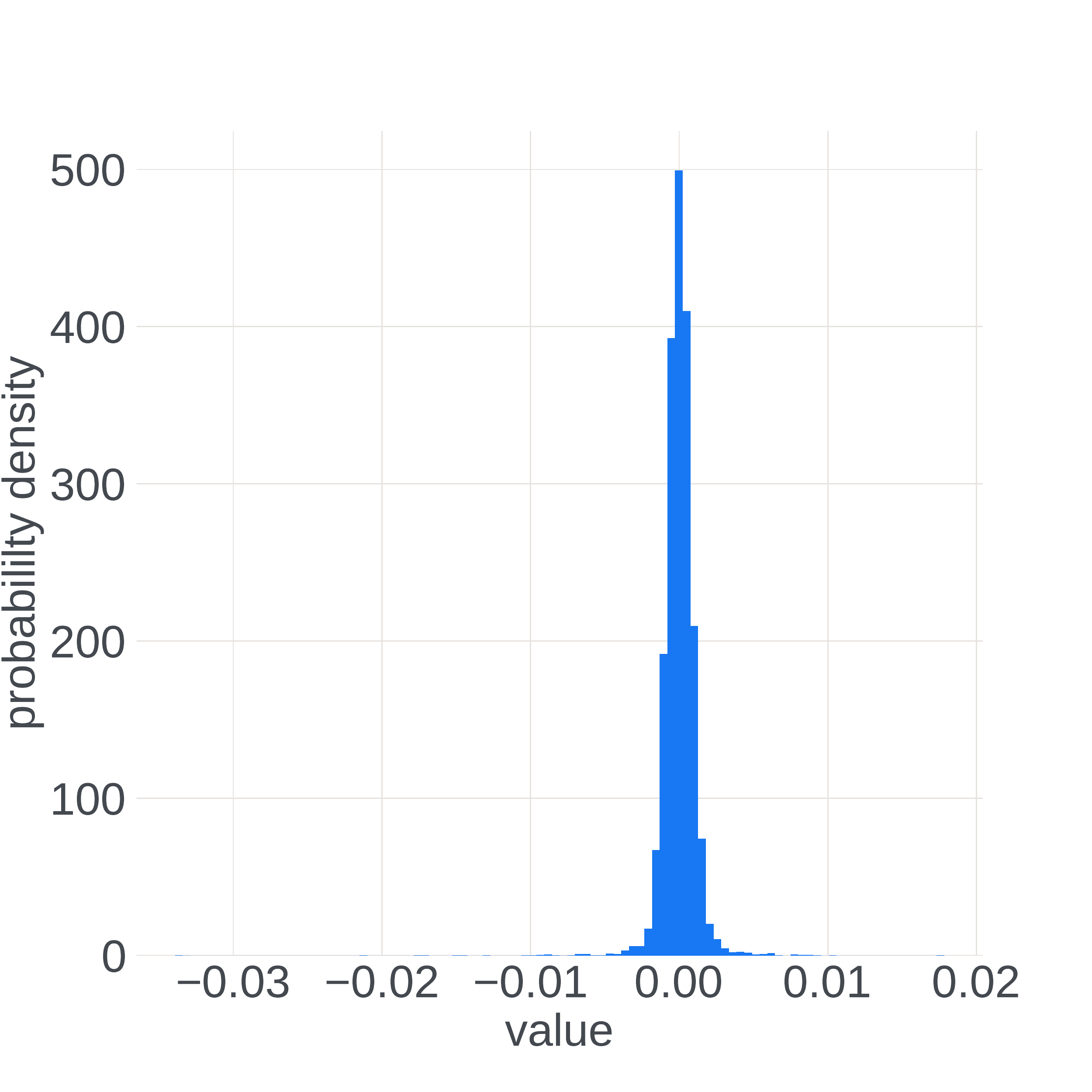}
         \caption{$65b$}
         \label{fig:spike_65b}
     \end{subfigure}
     \begin{subfigure}[b]{0.24\textwidth}
         \centering
         \includegraphics[width=\textwidth]{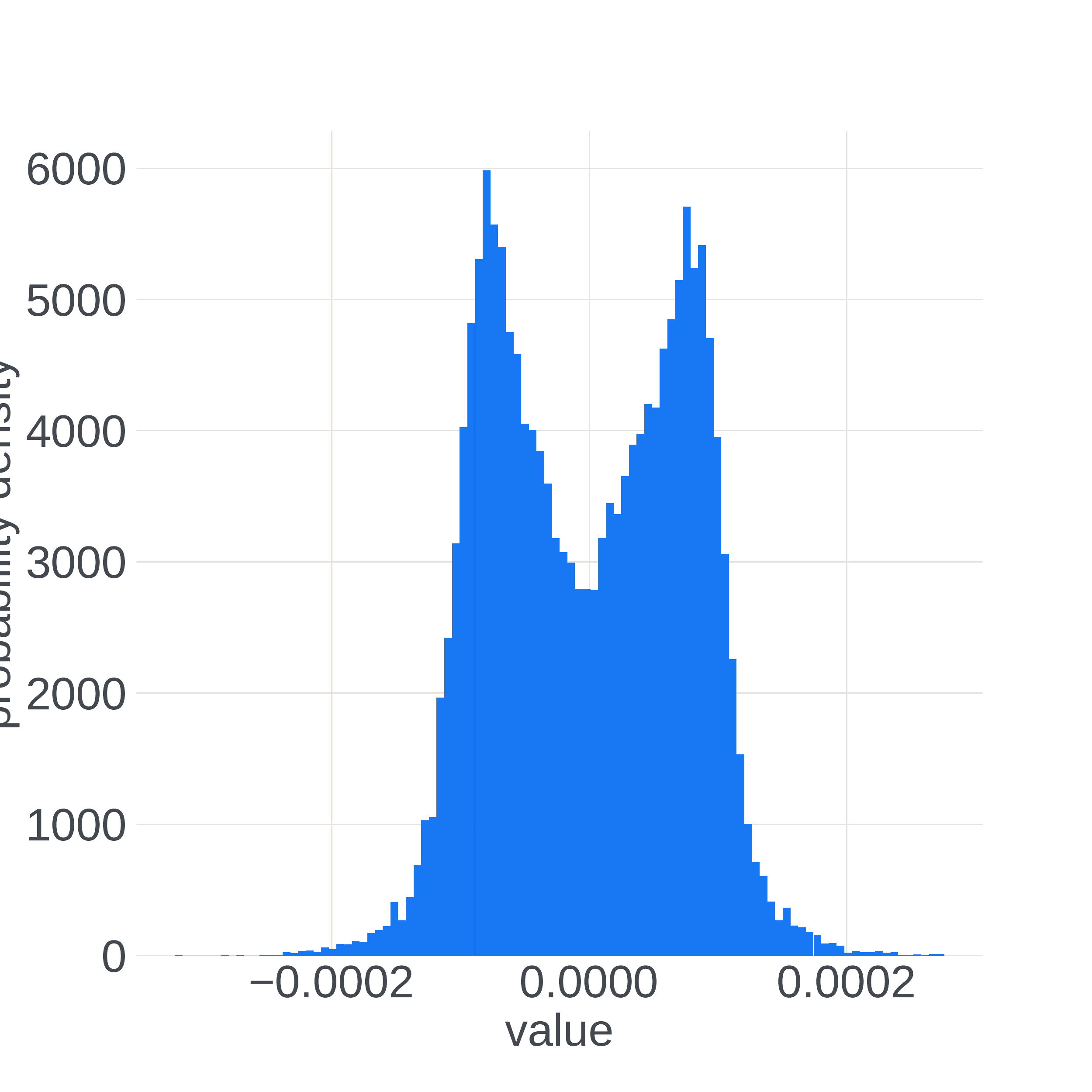}
         \caption{$546b$}
         \label{fig:spike_546b}
     \end{subfigure}
        \caption{Examples of the update $u[i, t]$ distribution over a layer $i\in G$ being concentrated around $0$ on various training steps $t.$ These distributions form spikes with the variance converging to zero with the growing size $n$ of the model.}
        \label{fig:spikes}
\end{figure}

Since the gradient estimation norm for the earlier layers of the model is comparable to $\varepsilon$ during normal operation, the approximation $\varepsilon \approx 0$ is not valid for those layers any longer. Thus, the space-domain distribution of the update components $u[G,t]$ over a layer $G$ ($i\in G$) with small gradient estimation norm must be different from the normal distribution predicted in Section \ref{sec:stat_indep} and from the bimodal distribution predicted in Section \ref{sec:stat_dep}. We should expect the distribution of $u[G, t]$ to concentrate around $0$ and take the form of a spike. The examples of the spiky distributions of the update components $u[G, t]$ for the layers of the models of different sizes can be seen in Figure \ref{fig:spikes}. Due to the gradient estimation components taking smaller values in larger models, the concentration effect is much more severe for the larger models than it is for the smaller ones.

\begin{figure}
     \centering
     \begin{subfigure}[b]{0.24\textwidth}
         \centering
         \includegraphics[width=\textwidth]{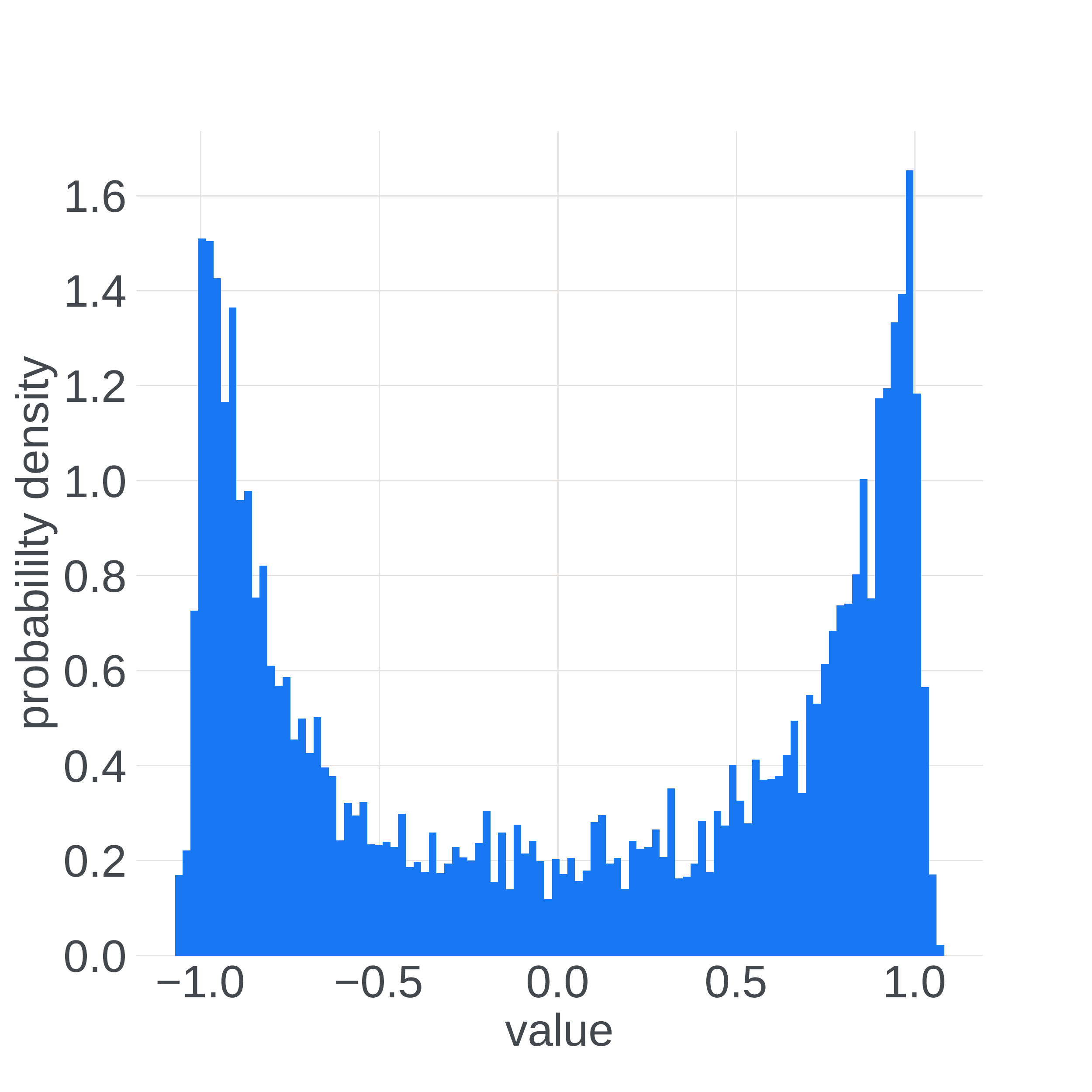}
         \caption{$6.7b$}
         \label{fig:ratio_7b}
     \end{subfigure}
     \hfill
     \begin{subfigure}[b]{0.24\textwidth}
         \centering
         \includegraphics[width=\textwidth]{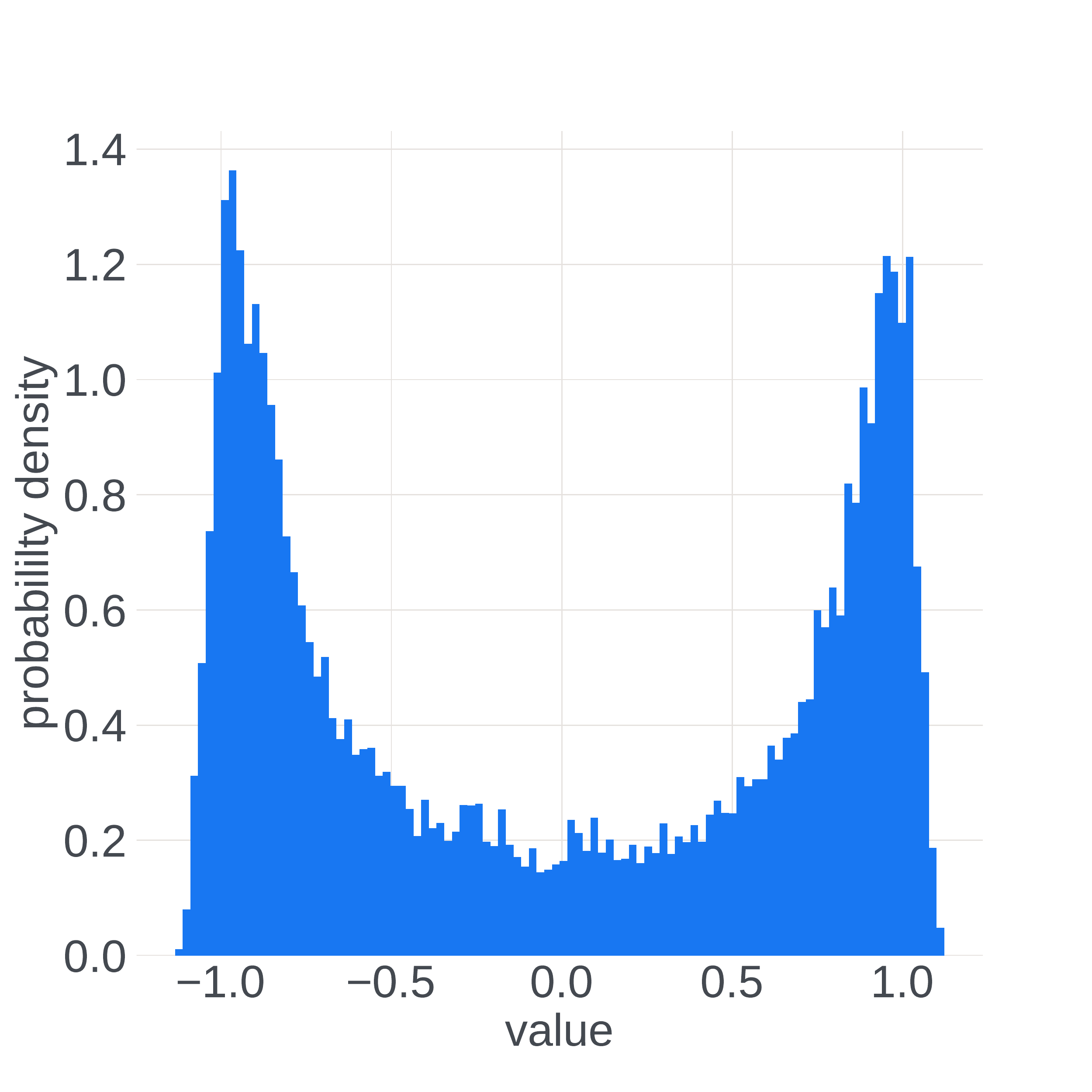}
         \caption{$30b$}
         \label{fig:ratio_30b}
     \end{subfigure}
     \hfill
     \begin{subfigure}[b]{0.24\textwidth}
         \centering
         \includegraphics[width=\textwidth]{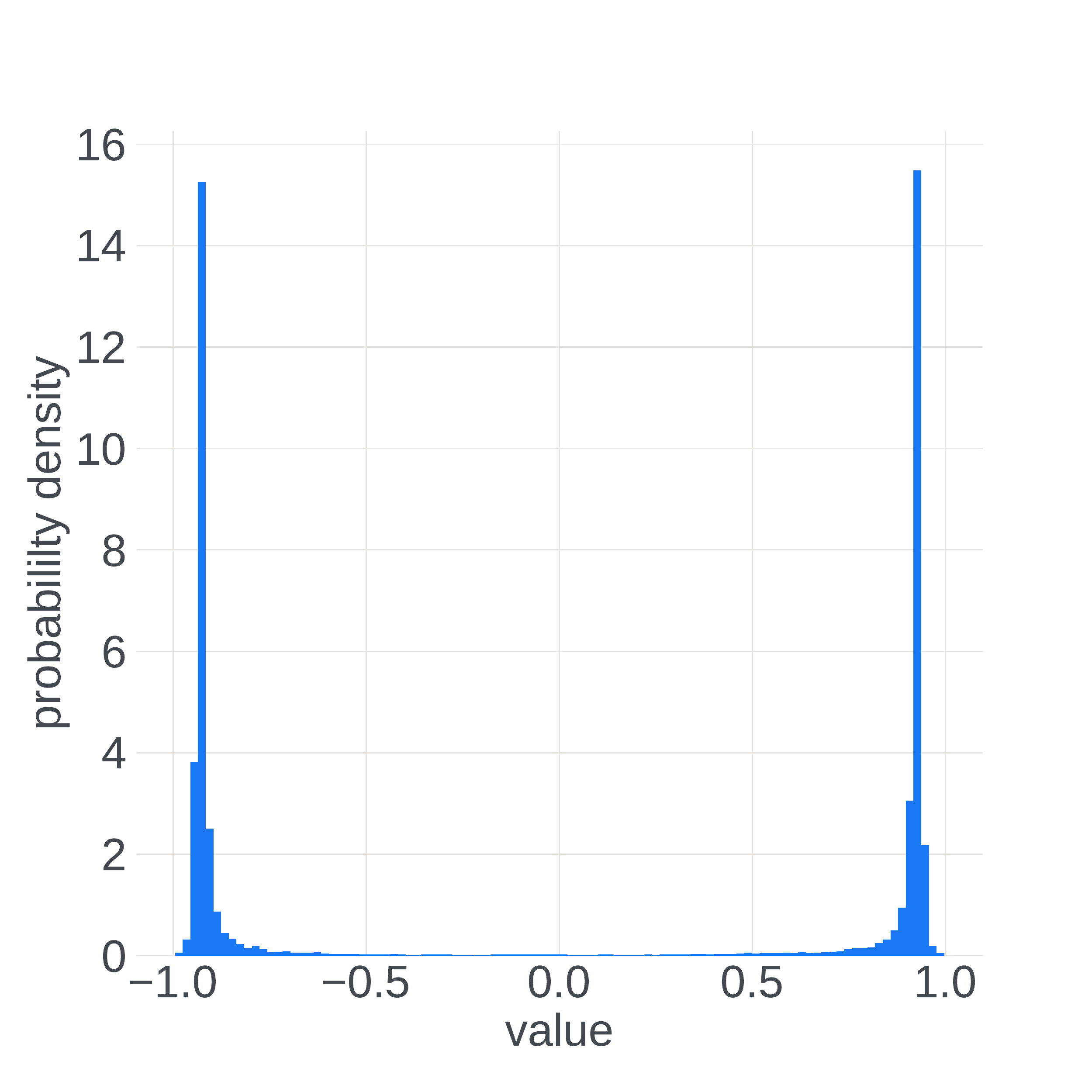}
         \caption{$65b$}
         \label{fig:ratio_65b}
     \end{subfigure}
     \begin{subfigure}[b]{0.24\textwidth}
         \centering
         \includegraphics[width=\textwidth]{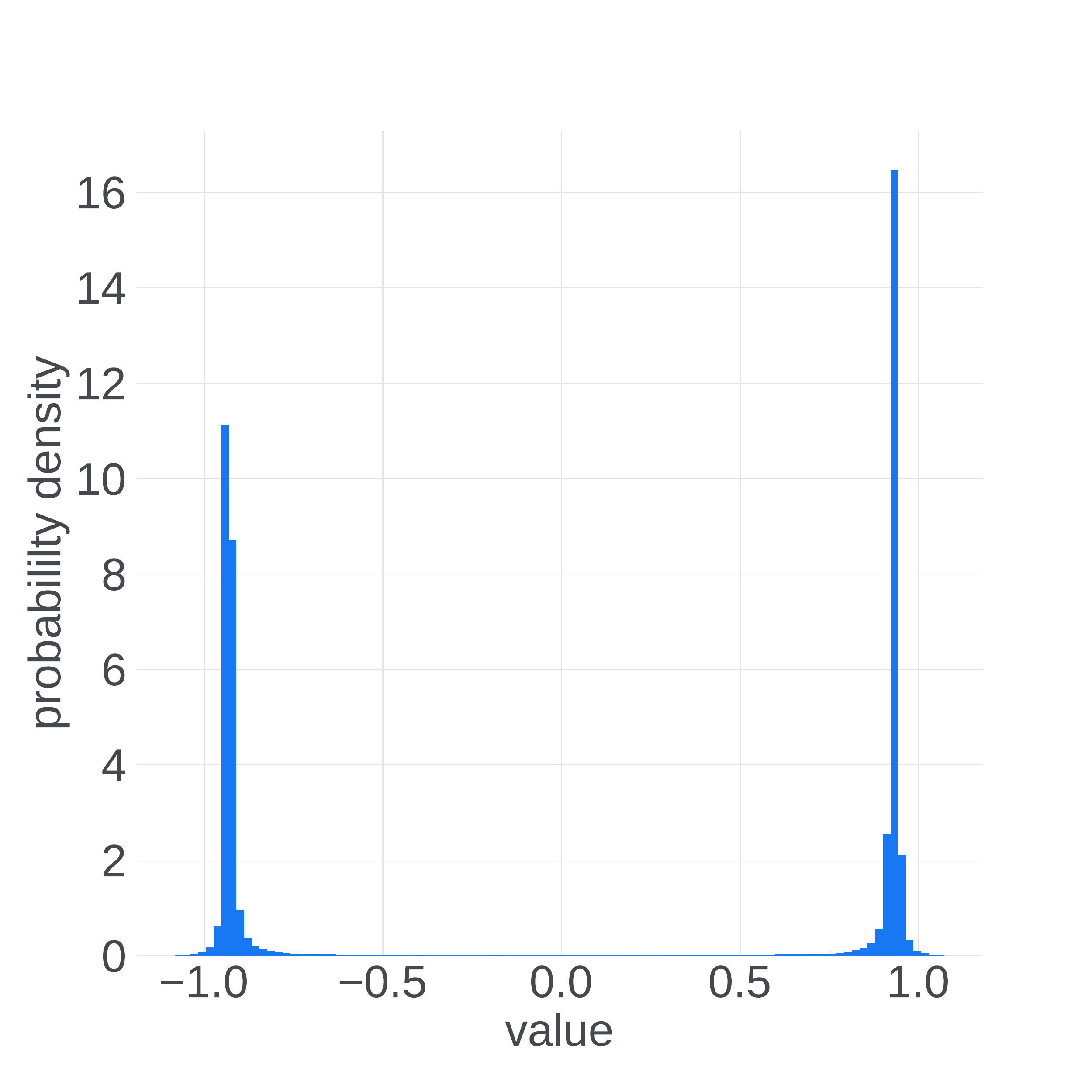}
         \caption{$546b$}
         \label{fig:ratio_546b}
     \end{subfigure}
        \caption{Examples of the ratio $r[G, t]=u[G,t] |_{\varepsilon=0}=\frac{m_t}{\sqrt{v_t}}$ distribution over a layer $i\in G$ at training steps $t$ which coincide with the layers and the training steps described in Figure \ref{fig:spikes}. The distributions take bimodal form, with each of the modes spiking at $\frac{\beta_1}{\sqrt{\beta_2}}\approx 0.92$ as the size $n$ of the model grows.}
        \label{fig:ratios}
\end{figure}

Let us separately consider the update $u[i, t] = \frac{m[i,t]}{\sqrt{v[i,t]} + \varepsilon}$ that depends on $\varepsilon$ and the ratio of the exponential moving average of the gradient and the exponential moving average of the square of the gradient (further just "ratio" for short): \[r[i,t]= \frac{m[i,t]}{\sqrt{v[i,t]}} =  u[i, t]\big |_{\varepsilon = 0}.\]
In Figure \ref{fig:ratios}, we plot the distributions of the ratio $r[G,t]$ of the same layers and at the same time steps as were used for Figure \ref{fig:spikes}. The distribution of $r[G,t]$ takes the bi-modal form discussed in Section \ref{sec:stat_dep}, and, unlike $u[G,t],$ is supported on an interval of order $1.$ The distributions of $u[G,t]$ and $r[G,t]$ will coincide as soon as the gradient estimation components scale back to the order above $\varepsilon.$ We conclude that, due to the large batch size and the vanishing values of the update components, the gradient estimation $g[i,t]$ for $i\in G$ have a strong time-domain correlation, thus the distribution of $r[i, t]$ is shifting to become bimodal. It is important to mention that in the early steps of training the distribution of $r[G,t]$ was noticed to have a bell shape, as discussed in Section \ref{sec:stat_indep}, even when the distribution of $u[G, t]$ is already spiked, although it is quickly degrading to the bimodal form further into training. This empirical evidence suggests that the distribution of the ratio remains the same even in the case of vanishing gradient estimation components.

\paragraph{Theory}

Based on the experimental and theoretical results, we propose the following explanation for the origins and the behavior of the training instabilities observed in large-scale machine learning. It takes the form of a multi-stage process spanning the short period of a single model perplexity spike:

\begin{enumerate}
    \item Healthy training: both $r[i, T]$ and $u[i, T]$ have uni-modal distributions, close to normal, with a standard deviation of the order of $1.$ This implies a low correlation between gradient estimates at consecutive steps and a high value of the gradient components relative to $\varepsilon$.
    \item The gradients of a group of parameters of the model (e.g. a layer, let us denote it with $G$) are vanishing ($\ll \varepsilon$) over the course of training. From our observations, this is most likely to happen in the earlier layers of the model, where a good feature representation has been learned, while the rest of the parameters further through the model keep gradients of high magnitude.
    \item The optimizer state values $m[i, t]$ and $v[i, t]$ for $i \in G$ vanishing ($\ll \varepsilon$).
    \begin{itemize}
        \item The update values $u[i, t]$ for $i \in G$ are vanishing due to $m[i, T]$ and $v[i, t]$ values dropping. The spatial distribution of $u[i, t]$ over $i\in G$ spikes at $0,$ dropping its variance. The distribution of $r[i, T]$ over $i\in G$ remains uni-modal for now, close to normal, variance of the order of $1$.
    \end{itemize}
    \item The distribution of $g[i, t]$ for $i \in G$ becomes highly correlated across time domain for two reasons:
    \begin{itemize}
        \item The model parameters remain unchanged ($\theta_{t+1} \approx \theta_t$) over the time steps as the update magnitude becomes close to zero.
        \item The batch size for a large model is usually also very large, thus the gradient evaluations have small time-domain variance.
    \end{itemize}
    \item The distribution of $r[i, t]$ over $i \in G$ changes from uni-modal to bi-modal. The distribution of $u[i, t]$ over $i\in G$ remains spiked at $0$.
    \item \label{item:rare_event} The model parameters $i$ outside the group $G$ slowly change their values because the values $u[i, t]$ corresponding to $i \not\in G$ are still of order of $1$. As the model changes, the probability that there would come a batch that would require a “reconsideration” of the feature maps learned in the earlier layers of the model, increases.
    Thus, increases the probability of a rare event (let's say it happens at time step $t^\star$), in which the $g[i, t^\star]$ for $i \in G$ becomes larger than $\varepsilon$. After this event, the distribution of $u[i, t]$ is going to depart from the spike form, approaching the bimodal distribution of $r[i, t]$. According to our analytical study described in Section \ref{sec:alignment}, this implies divergence for the standard learning rate values, which means that the entries of the next gradient estimation $g[G, t^\star+1]$ must get an even larger magnitude, bringing the distribution of $u[i, t^\star+1]$ over $G$ even closer to the distribution of $r[i, t^\star+1]$. The process described here resembles a chain reaction, which we would expect to observe during a loss explosion. In Figures \ref{fig:heatmaps} it could be observed that the explosion of the gradient norm in the earlier layers of the model starts one training step earlier than the spike of perplexity metric.
    \begin{itemize}
        \item Note that for a function of the form $\phi(x, \varepsilon) = \frac{x}{|x|+\varepsilon}$, the derivative $\frac{\partial \phi}{\partial x}\big|_{x=0} = 1/\varepsilon$ is large for small values of $\varepsilon$, which means that low changes in the gradient estimation lead to a disproportionately large change in the Adam update magnitude.
    \end{itemize}
    \item The spatial distribution of $u[i, T]$ over $i \in G$ approaches the bimodal distribution of $r[i, t]$, leading to divergent behavior as discussed earlier.
    \item The gradient estimation components $g[i, t]$ for $i \in G$ start to vary widely over large magnitudes, losing correlation in the time domain. Thus, the spatial distribution of $r[i, t]$ becomes uni-modal. Since $g[i, t] \gg \varepsilon$, the distribution of $u[i, t]$ coincides with the one of $r[i, t]$ becoming uni-modal as well.
    \item The training becomes healthy again. If the size of the group $G$ is relatively small, the loss drops back to pre-explosion value very quickly, learning the good features again. However, it has been observed that the loss does not always come back to the pre-explosion values, leading to divergence.
\end{enumerate}

The description of the process will inform our discussion of the possible remedies for the training instabilities provided in Section \ref{sec:discussion}.

\subsection{Snapshots of the large model training run in the unstable regime}

\begin{figure}
     \centering
     \begin{subfigure}[b]{0.4\textwidth}
         \centering
         \includegraphics[width=\textwidth]{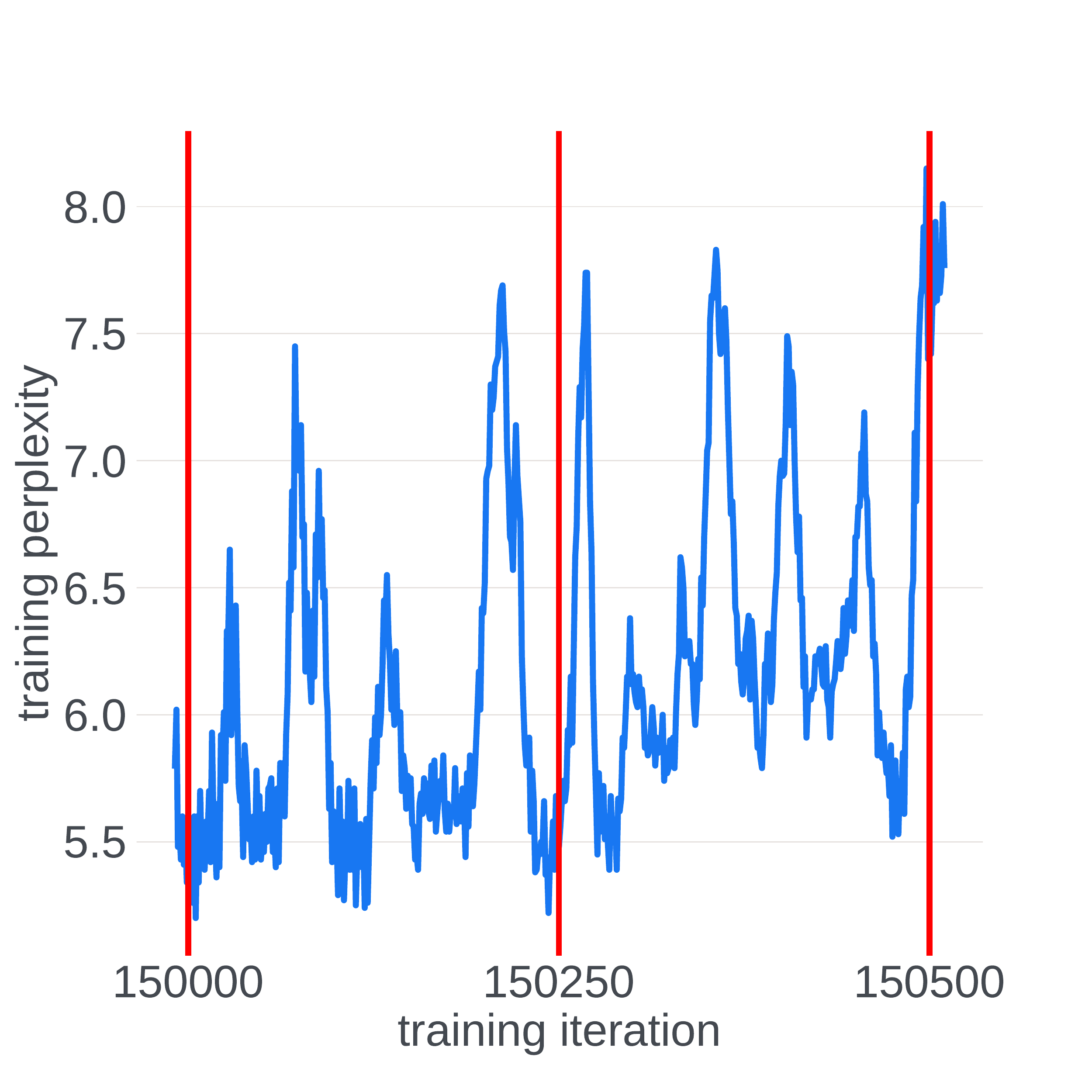}
     \end{subfigure}
        \caption{Sudden spiky behavior of the training perplexity curve of a large-scale training run. The training iterations $t_1=150000,$ $t_2=150250$ and $t_1=150500$ are marked with red vertical stripes.}
        \label{fig:three_ppl}
\end{figure}

\begin{figure}
     \centering
     \begin{subfigure}[b]{0.3\textwidth}
         \centering
         \includegraphics[width=\textwidth]{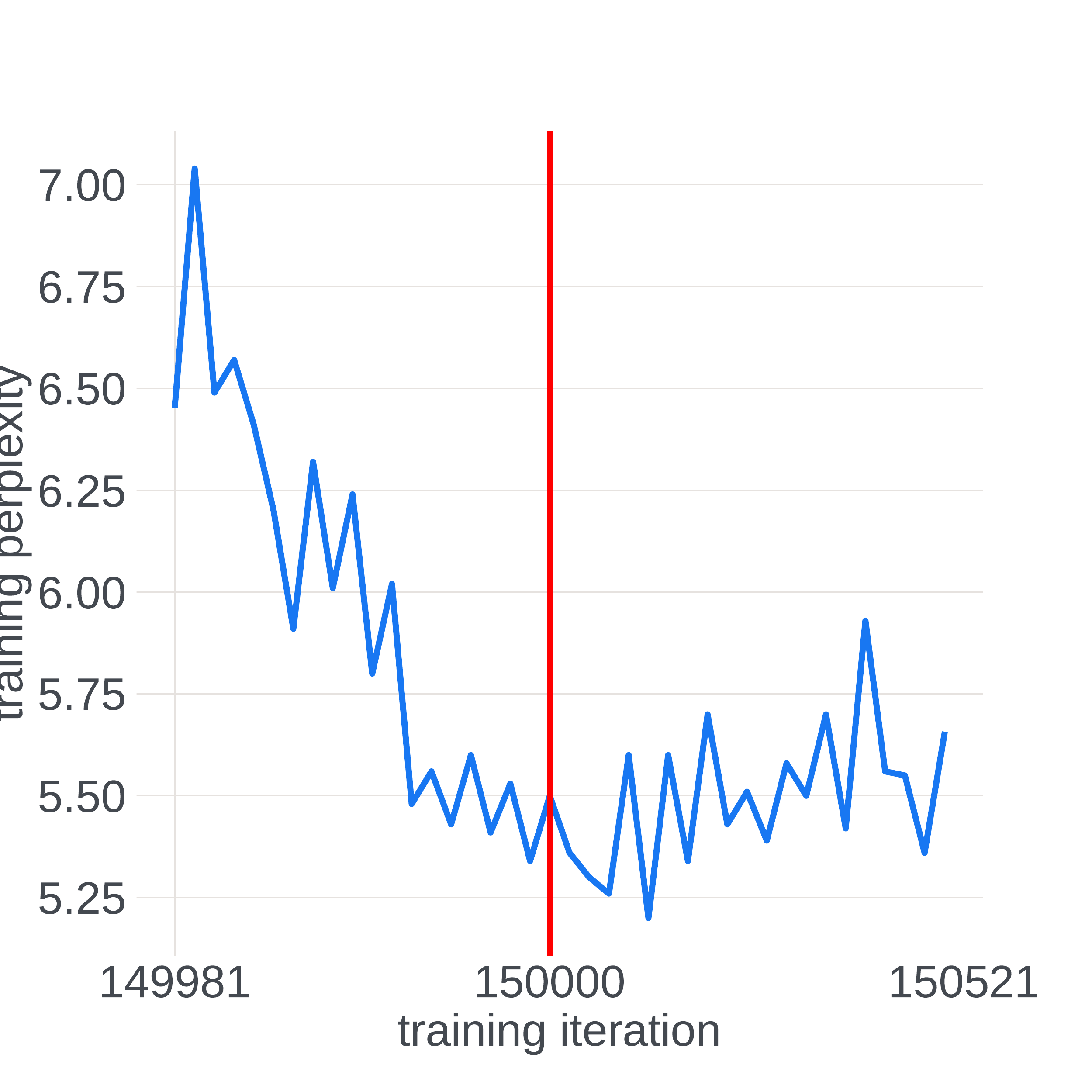}
         \caption{perplexity curve around $t_1$}
         \label{fig:550b_ppl_1}
     \end{subfigure}
     \hfill
     \begin{subfigure}[b]{0.3\textwidth}
         \centering
         \includegraphics[width=\textwidth]{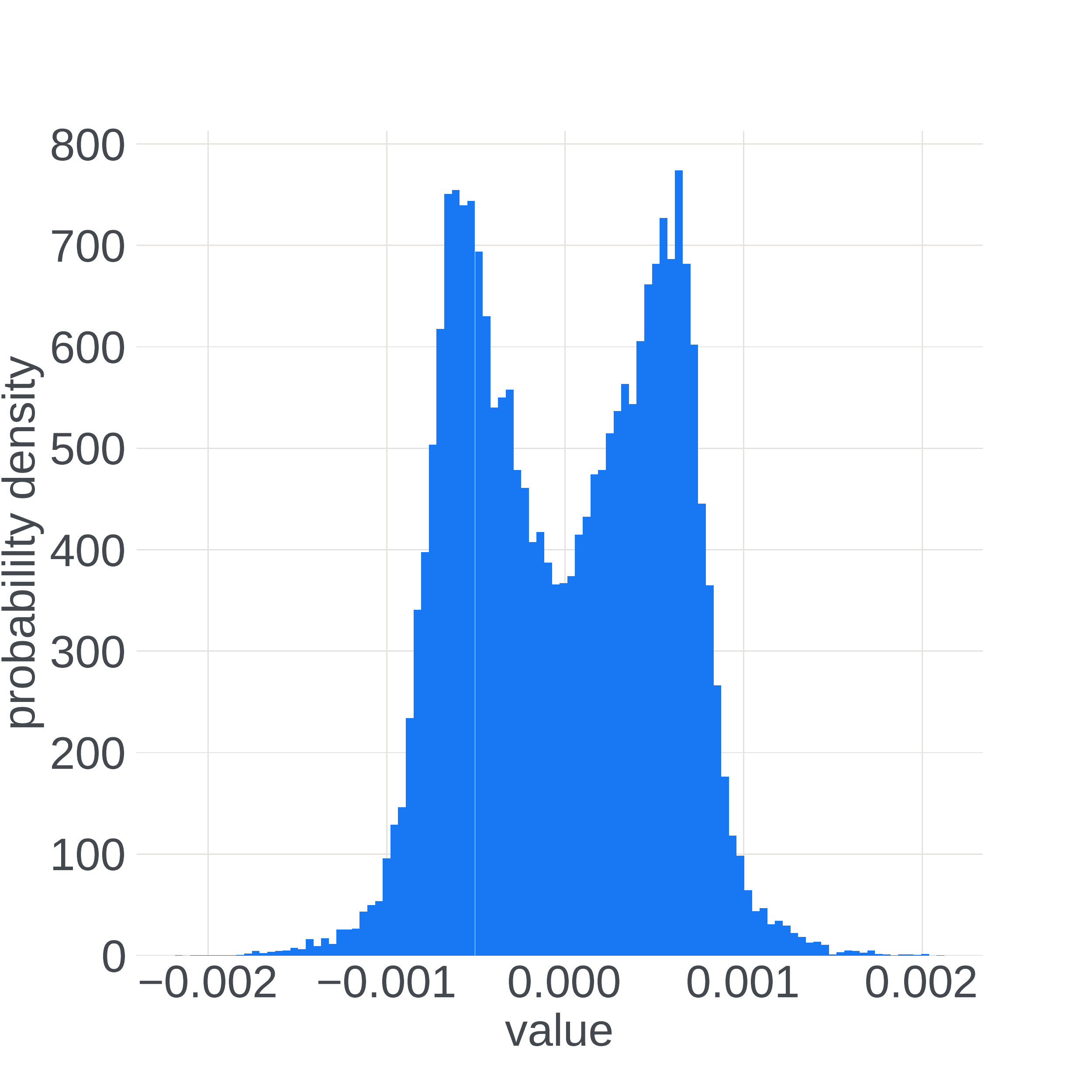}
         \caption{$u[G_0, t_1]$}
         \label{fig:msq_546b_u1}
     \end{subfigure}
     \hfill
     \begin{subfigure}[b]{0.3\textwidth}
         \centering
         \includegraphics[width=\textwidth]{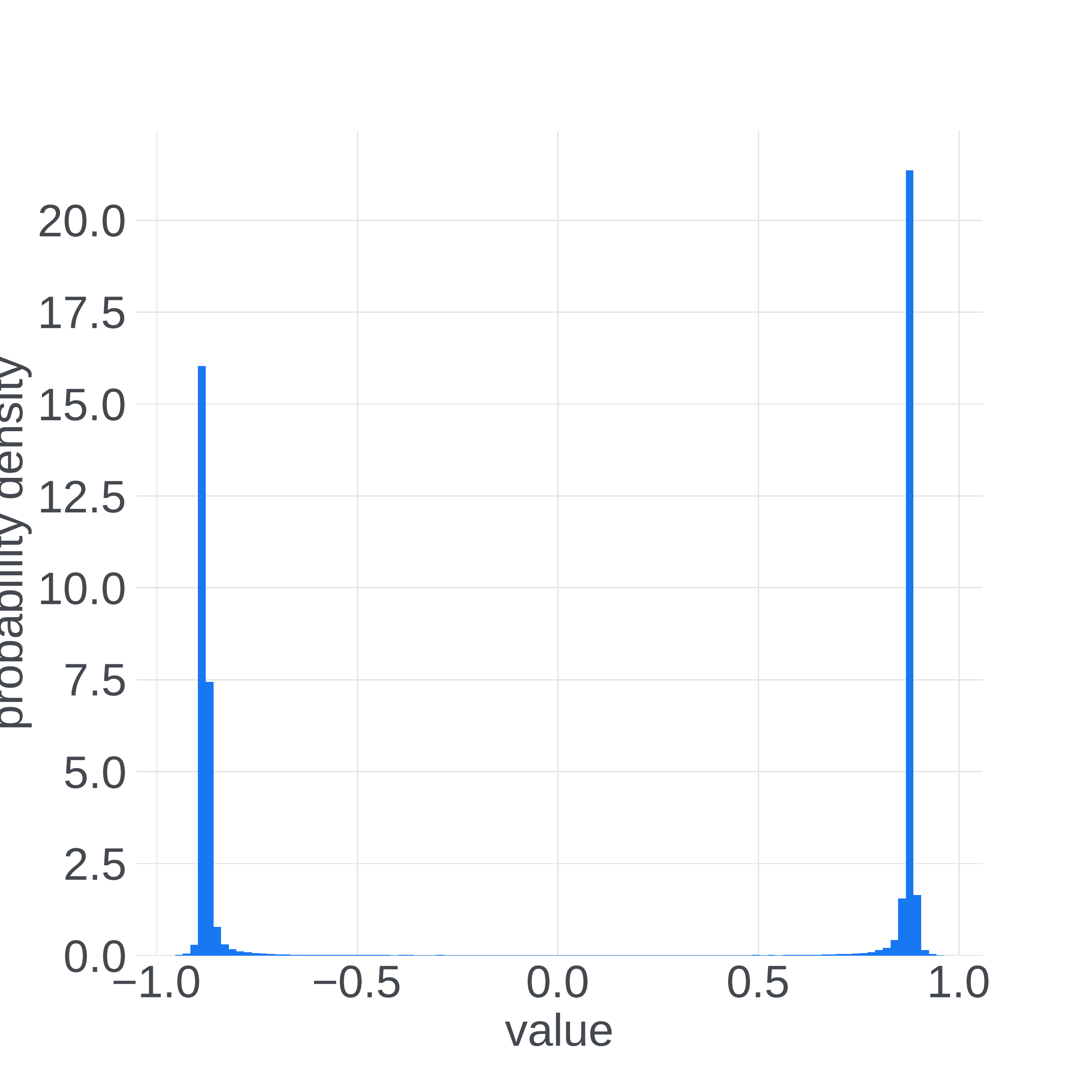}
         \caption{ $r[G_0, t_1]$}
         \label{fig:msq_546b_r1}
     \end{subfigure}
        \caption{Distributions of $u[i, t_1]$ and $r[i, t_1]$ over the parameters $i$ belonging to the token embedding layer $G_0$ of the $546$b model at the time step $t_1=150000,$ and the training perplexity curve for the iterations around $t_1.$ }
        \label{fig:iteration1}
\end{figure}

\begin{figure}
     \centering
     \begin{subfigure}[b]{0.3\textwidth}
         \centering
         \includegraphics[width=\textwidth]{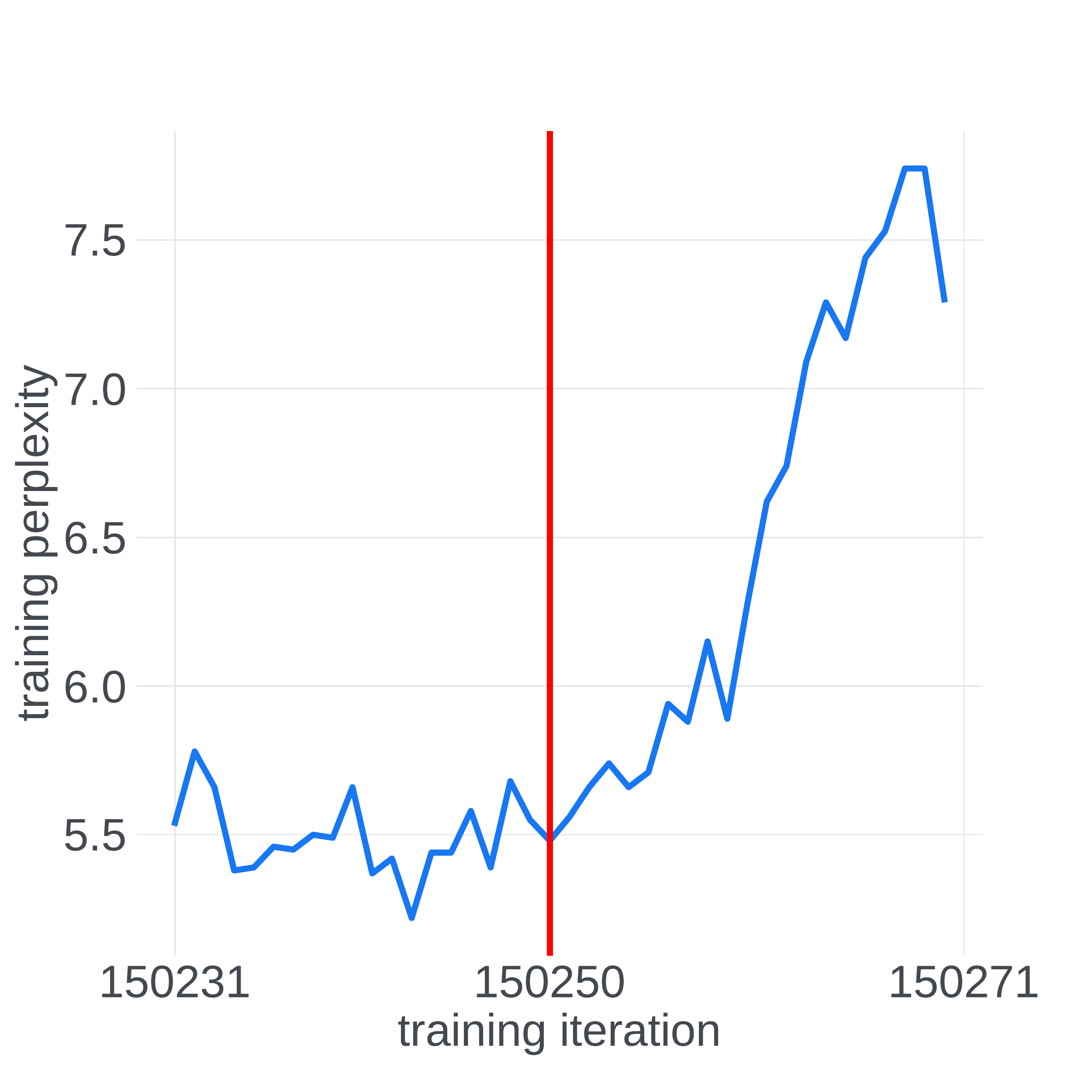}
         \caption{perplexity curve around $t_2$}
         \label{fig:550b_ppl_2}
     \end{subfigure}
     \hfill
     \begin{subfigure}[b]{0.3\textwidth}
         \centering
         \includegraphics[width=\textwidth]{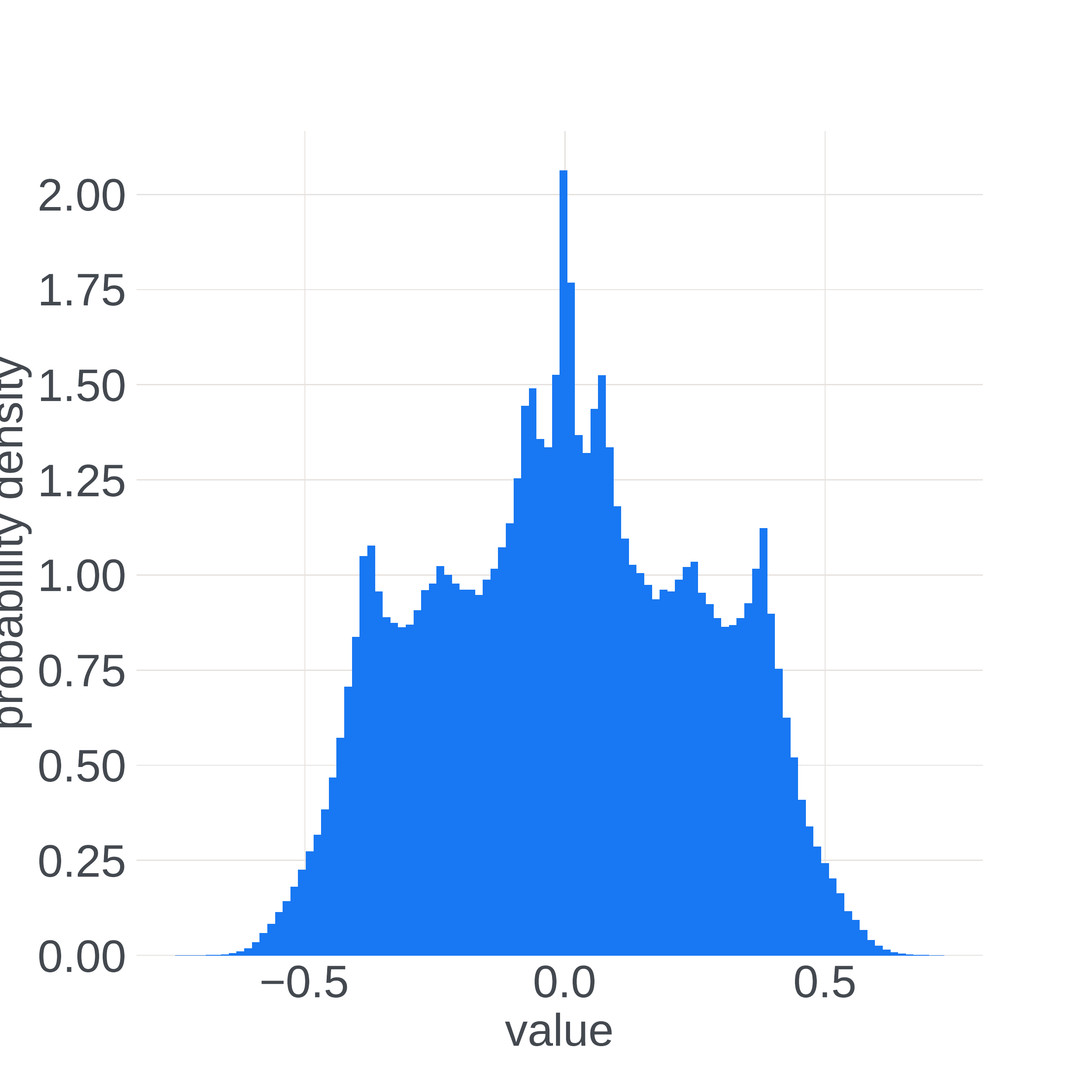}
         \caption{$u[G_0, t_2]$}
         \label{fig:msq_546b_u2}
     \end{subfigure}
     \hfill
     \begin{subfigure}[b]{0.3\textwidth}
         \centering
         \includegraphics[width=\textwidth]{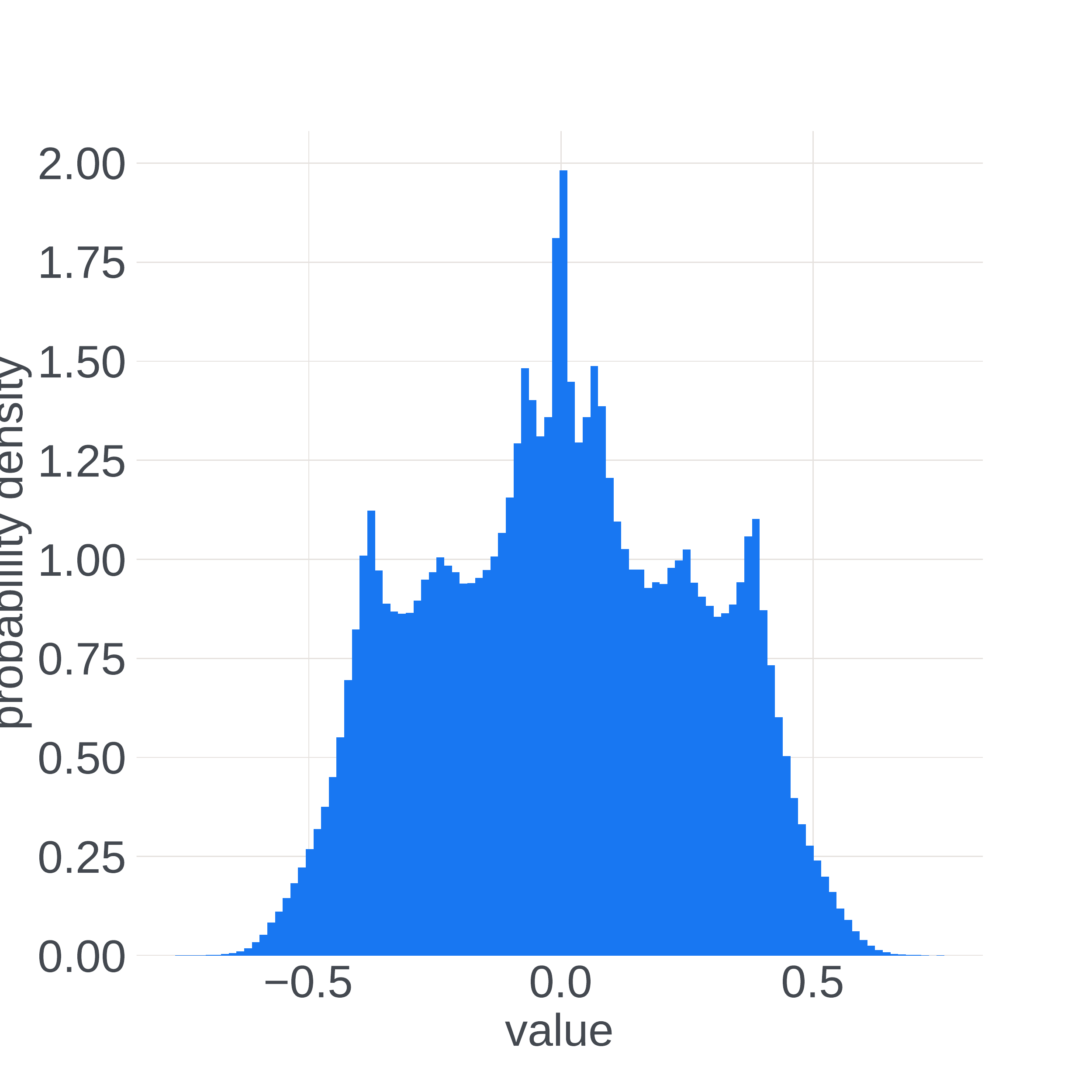}
         \caption{$r[G_0, t_2]$}
         \label{fig:msq_546b_r2}
     \end{subfigure}
        \caption{Distributions of $u[i, t_2]$ and $r[i, t_2]$ over the parameters $i$ belonging to the token embedding layer $G_0$ of the $546$b model at the time step $t_2=150250,$ and the training perplexity curve for the iterations around $t_2.$  }
        \label{fig:iteration2}
\end{figure}

\begin{figure}
     \centering
     \begin{subfigure}[b]{0.3\textwidth}
         \centering
         \includegraphics[width=\textwidth]{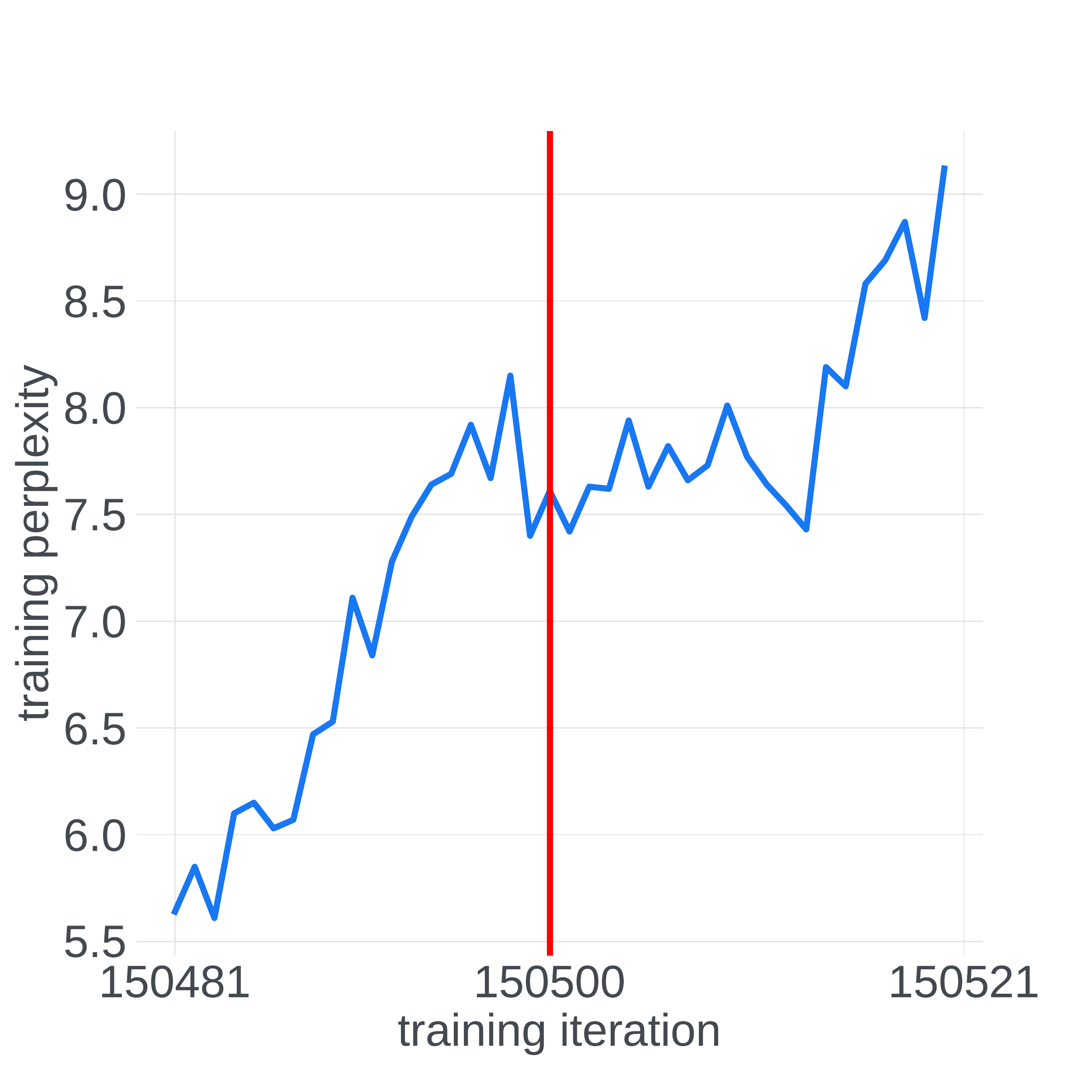}
         \caption{perplexity curve around $t_3$}
         \label{fig:550b_ppl_3}
     \end{subfigure}
     \hfill
     \begin{subfigure}[b]{0.3\textwidth}
         \centering
         \includegraphics[width=\textwidth]{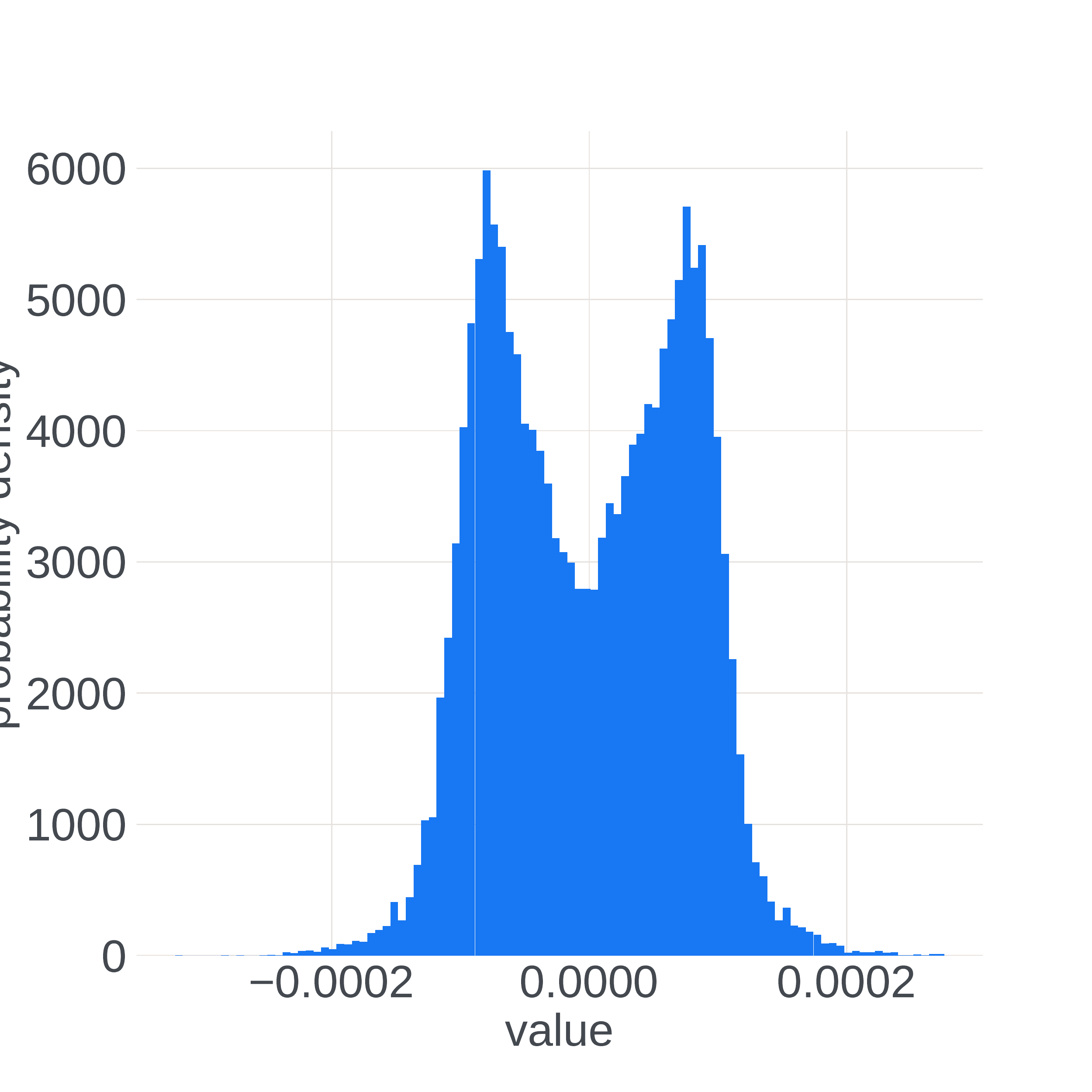}
         \caption{$u[G_0, t_3]$}
         \label{fig:msq_546b_u3}
     \end{subfigure}
     \hfill
     \begin{subfigure}[b]{0.3\textwidth}
         \centering
         \includegraphics[width=\textwidth]{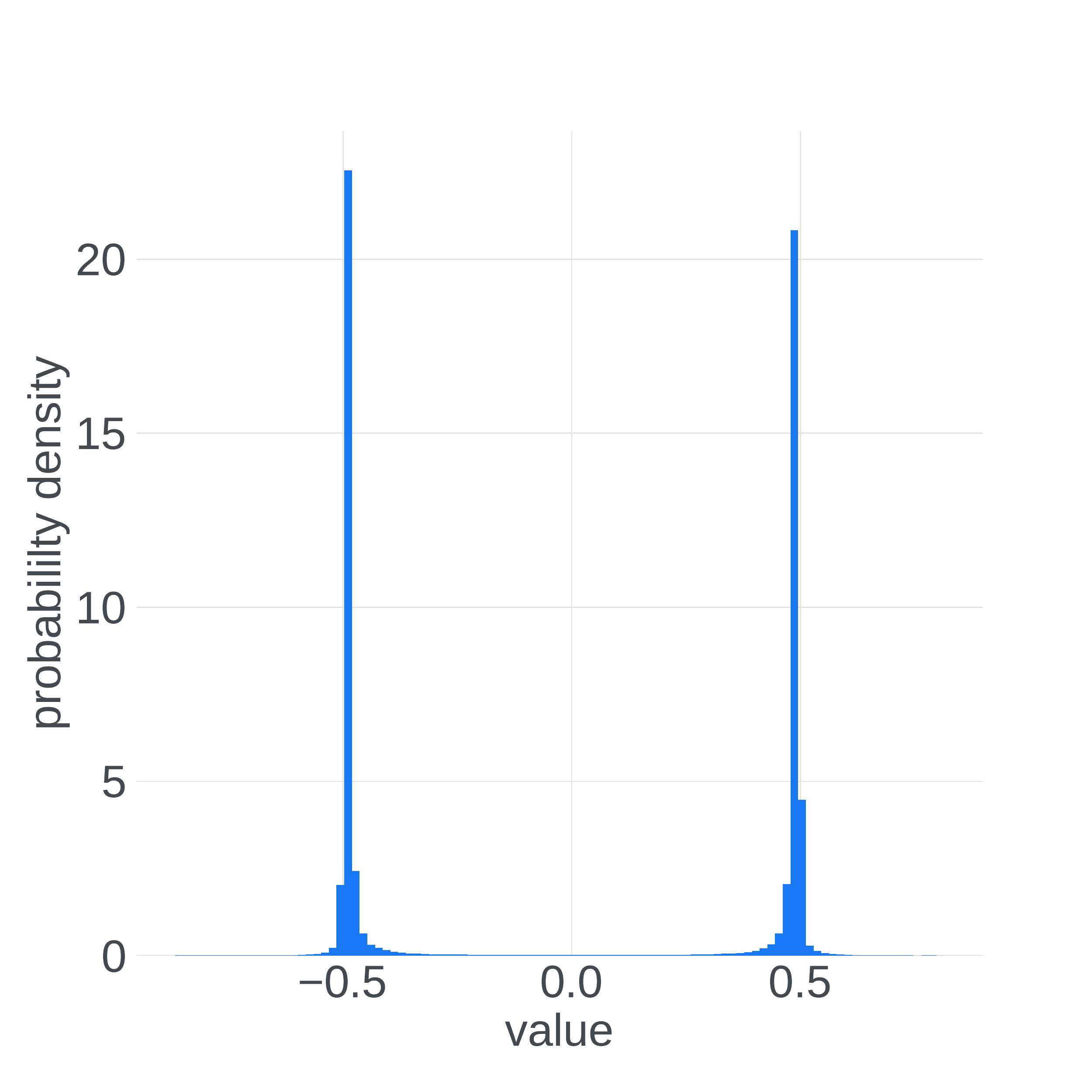}
         \caption{$r[G_0, t_3]$}
         \label{fig:msq_546b_r3}
     \end{subfigure}
        \caption{Distributions of $u[i, t_3]$ and $r[i, t_3]$ over the parameters $i$ belonging to the token embedding layer $G_0$ of the $546$b model at the time step $t_3=150500,$ and the training perplexity curve for the iterations around $t_3.$ }
        \label{fig:iteration3}
\end{figure}

In this section, we discuss the snapshots of the $546$b model taken at the final stages of the largest training run we report, where the instabilities are the most prominent feature of the training. In Figure \ref{fig:three_ppl} we plot the learning curve throughout approximately $500$ training iterations. The three distinct training iterations to which we refer as $t_1 = 150000,$ $t_2=150250,$ and $t_3=150500$ are marked with red lines. For all three of the iterations, we demonstrate the behavior of the training perplexity value in the proximity of the iteration as well as the distributions of the update $u[G, t]$ and the ratio $r[G,t] = u[G,t]\big|_{\varepsilon = 0}$ components over the embedding layer $G_0$ of the network. Although the majority of layers did not experience a special state at the time steps under consideration, there were other layers in the model that showed a multi-modal behavior of their update distributions at the same training steps. We demonstrate the first layer as a good representative of them. All of the examples are given in Figures \ref{fig:iteration1}, \ref{fig:iteration2}, and \ref{fig:iteration3}. Each of the iterations considered here represents a stage of the loss explosion process. $t_2$ is an iteration where the loss explosion has started, $t_1$ is an iteration at which the loss has returned to an approximately pre-explosion value and $t_3$ is an iteration on which the loss explosion has ended but went flat instead of dropping back down.

It is important to note that Figures \ref{fig:msq_546b_u1} and \ref{fig:msq_546b_r1} and resemble Figures \ref{fig:msq_546b_u2} and \ref{fig:msq_546b_r2} which are associated with two different phases of a perplexity spike.
This highlights the question of when the loss drops back down and when it does not, which remains outside of the scope of the proposed theory. We can hypothesize here that the feature representation of the earlier layers changes more drastically at times when the drop of the loss does not occur than in the cases where the drop happens. As expected at $t_2,$ the distribution of $u[G, t_2]$ closely resembles the distribution of $r[G, t_2],$ although both of these distributions have come a long way to meet from the state depicted in Figures \ref{fig:msq_546b_u1} and \ref{fig:msq_546b_r1}.

\section{Discussion}\label{sec:discussion}

In this section, we summarize the observations that we have considered to falsify our theory or generally found relevant to the phenomenon being under study. We found it interesting that the distribution of the components of the ratio $r[:,t]$ can turn from bimodal to a uni-modal and back without a prominent spike in the model perplexity, which is happening due to a slight increase in the magnitude of gradient estimation components which do not surpass the value of $\varepsilon.$ Thus, the jump in the norm of gradient estimation does not always imply a loss spike, which would be the case if the distributions of $r[G,t]$ were mostly uni-modal at the time of the spike for all of the layers $G.$

The theory already explains quite well how the model loss can return to the pre-explosion values shortly after the instability begins. However, it is also critically important for the proposed theory to explain why the trick with skipping batches helps for the stabilization of the training loss curve. We can see that step \ref{item:rare_event} above requires a rare event to happen, in which the order of magnitude of the gradient estimation components for a particular part of the model parameters experiences a sudden jump above the value of $\varepsilon.$ Thus, skipping the batch that brought a large gradient value means skipping this rare event that would otherwise start a chain reaction because of the state of the optimizer having a severely bimodal distribution of $r[:,t].$ Taking a gradient step through the model where the optimizer is in the state with $r[:,t]$ having a bell shape distribution does not lead to the chain reaction even if it provokes a spike in the norm of gradient estimation.

While in the smaller or shallower models, the distribution of the update components is close to normal with the variance of order one, the deeper and larger models develop layers with the update components distribution spiked at zero rather quickly. It is much more common for decoder-only transformers to develop a small magnitude of the gradient estimates for the layers situated earlier in the network. This feature is likely related to the classical problem of vanishing gradients \citep{hochreiter2001gradient}, although as could be seen in Figure \ref{fig:heatmaps}, the vanishing effect is not monotone in the layer number. Our theory draws a connection from the loss instability to the problem of diminishing gradients and to the batch size used for training. Thus, the size of the models, which is tightly bound to the model depth and the batch size, should indeed be the crucial hyper-parameter that strongly correlates with how prone the training is to the observed instabilities.




\subsection{How can the loss spikes be mitigated?}

We point out several possible ways to fight the training instabilities caused by the time-domain correlation of gradient estimations. Each of them has its limitations and we discuss the trade-offs in detail to inform the future experiment design.

As proposed by \citet{chowdhery2022palm}, skipping batches is a possible strategy for recovering the loss curve, although it only works well earlier in training. Skipping batches is challenging to implement and operate, as it usually requires manual intervention to keep track of training loss.
It is consuming resources that are spent on rolling back the model and saving checkpoints more frequently. Most importantly, the frequency of the spikes increases later in training when the order of magnitude of the gradient estimation entries for the majority of layers drops below $\varepsilon.$ The situation when the instability of the training loss becomes unbearable is illustrated in Figure \ref{fig:three_ppl}.

Alternative options include lowering the learning rate, which helps both theoretically and in practice but extends the overall training time. Tuning down the $\varepsilon$ value should be considered, but it does not go well together with the training in low-precision arithmetic which is popular in large-scale training due to its efficient use of inter-GPU communication bandwidth. Changing the approach to treating the division by zero might be necessary. For example, in the process of studying the ratio values $r[i,t] = u[i,t]\big|_{\varepsilon=0}$ we noticed that the order of magnitude of the ratio components never exceed the order of $1$ and the only cases when "Not a Number" values are observed are when both $m_t[i]$ and $v_t[i]$ are equal to $0$. No cases of $m_t[i]\ne v_t[i] = 0$ have been observed. Thus, putting $\varepsilon=0$ and mapping $u[i,t]$ to $0$ in case $v_t[i]=0$ may be one option to avoid numerical issues and prevent the bi-modal distribution of the ratio from forming.
We expect the reduction of batch size to help reduce the frequency of loss spikes due to increased variance in the gradient estimation, but it would also slow down the large-scale training that exploits data parallel distribution paradigm over a large number of machines, as the batch size per GPU might be too small to make use of the GPU acceleration of tensor operations.

Reducing the averaging constants $\beta_1$ and $\beta_2$ would lead to averaging gradients over a larger number of time steps and would increase the time frame over which the correlation between the gradient estimations would need to persist for the severe bi-modality to appear in the distribution of the update components. On the downside, this would make the update stale with not enough up-to-date information about the gradient during the normal phase of training.

It was observed in our experiments on a smaller scale that the composition of the training dataset can reduce the number of layers with the diminishing gradient values down to zero. For the natural language processing tasks, this happened when the text corpus consisted of high-quality data with a lower variety of text modalities.

A conceptually different way to take care of training instabilities would be to keep track of a statistic that measures uni-modality of the distribution of the ratio $r_t = \frac{m_t}{\sqrt{v_t}}$, and tune down the $\varepsilon$ value, or even completely reinitialize the optimizer state, whenever the distribution changes its shape. One example of such a statistic is the dip statistic proposed by \citet{hartigan1985dip}. Initial experiments in high-precision training have shown that this strategy allows preventing the bi-modal distribution of the updates from forming.

\section{Related Work}\label{sec:background}

In this Section, we briefly discuss (a not exhaustive) list of prior works on instabilities and divergence of the Adam algorithm. We focus on the questions raised in these papers and their main contributions to highlight why none of the results of the prior studies could explain the phenomenon of instability of the large-scale training runs.

There is a series of works refuting the initial claims of Adam convergence, setting limits for its applicability.
\citet{reddi2019convergence} pointed out an issue with the theoretical convergence of Adam even in a low-dimensional setting. They considered the Adam update to be the component-wise product of the average gradient and a vector of "learning rates". They notice that each component of the vector of learning rates does not monotonically decrease throughout optimization steps. Using this observation, they come up with examples of the divergent behavior of Adam and propose alternative algorithms called AMSGrad and AdamNC.
\citet{wang2022divergence} propose another simple one-dimensional unconstrained problem of population loss minimization that illustrates the divergent behavior of Adam. The example shows the model drifts away from the optimal solution over the course of iteration, despite the strong convexity of the objective and the absence of constraints. They identify the variance of the gradient estimates as the key concern and propose a variance-reduced Adam algorithm that manages to converge under certain assumptions. This line of work is aimed at a different phenomenon, namely the drifting divergence, rather than the rapid spiky divergence we are concerned with.

\citet{chen2018convergence} conducted an analysis that involved comparing the direction of the gradient and the direction of the update at each step of Adam dynamics, which is an idea we exploit as well. Their study resulted in conditions on certain characteristic dynamic quantities, sufficient for the convergence of the Adam algorithm. The quantities could theoretically be monitored but the conditions can not be easily imposed on a training problem in practice. We did not follow the suggested quantities throughout the training run in our experiments, although it may give additional insights into the cause of the spike.
Another sufficient condition on the convergence of Adam-like algorithms was presented by \citet{zou2019sufficient}. Their conditions are set on the hyper-parameters of the algorithms, rather than dynamic quantities. They establish convergence rate guarantees for Adam and conclude that these rates do not hold for the commonly used version of the Adam algorithm due to hyper-parameter values. The issues discussed by the authors are also more relevant to explaining the drifting divergence phenomenon. Another work aimed at drifting divergence was done by \citet{zaheer2018adaptive} who proposed a new method, Yogi, motivated by the concern that the dependence of Adam updates on the past gradient estimations decay exponentially with the number of steps.

\citet{zhou2018adashift} argue that the fundamental reason behind Adam divergence is the negative correlation between the scale of gradient estimation and the vector of learning rates, which results in a small step size for a large gradient, and a large step size for a small gradient. This reasoning does not agree with our observations, as the distribution of the Adam update was repeatedly seen to have the same shape independently of the scale of the gradient estimation (given that the gradient estimation is much larger than the stability constant $\varepsilon$).

\section{Conclusion}\label{sec:conclusion}

In this work, we argue that the training loss instabilities observed in large-scale training should be associated with the time-domain correlation between the gradient estimates of earlier layers in the deep-learning models. Based on the identified connection, we propose several ways to mitigate the instabilities, along with the heuristic method that was known in the literature. We conclude that at this point, there is no silver bullet to solve the problem, and the appropriate remedy depends on the specific setup of the large-scale training run.

\section*{Acknowledgement}
The authors are grateful to Sho Yaida for his time spent carefully reading and commenting on the paper.

\bibliography{output}

\end{document}